\documentclass[runningheads]{llncs}

\usepackage[review,year=2026,ID=5630]{eccv}

\usepackage{eccvabbrv}

\newif\ifarxiv
\arxivtrue

\makeatletter
\@ifundefined{ifarxiv}{\newif\ifarxiv}{}   
\makeatother

\newcommand{\supplref}[1]{\ifarxiv Section~\ref{#1}\else the Supplementary\fi}

\newcommand{\supplorinline}[2]{\ifarxiv #2\else #1\fi}

\usepackage{booktabs}
\usepackage{multirow}
\usepackage{makecell}
\usepackage{tcolorbox}
\usepackage{fvextra}
\usepackage{marginnote}
\usepackage{wrapfig}

\DeclareRobustCommand{\tok}[1]{\texttt{\detokenize{<|#1|>}}}

\newcommand{\rad}{\textup{\textsc{RAD}}\xspace}
\newcommand{\lad}{\textup{\textsc{LAD}}\xspace}

\definecolor{eccvblue}{rgb}{0.12,0.49,0.85}
\usepackage[pagebackref,breaklinks,colorlinks,citecolor=eccvblue]{hyperref}

\usepackage[accsupp]{axessibility}

\title{\rad-\lad: Rule and Language Grounded Autonomous Driving in Real-Time}
\titlerunning{RAD-LAD}

\author{Anurag Ghosh\inst{1} \and
Srinivasa Narasimhan\inst{1} \and
Manmohan Chandraker\inst{2,3} \and
Francesco Pittaluga\inst{2}
}
\authorrunning{A.~Ghosh et al.}
\institute{$^{1}$Carnegie Mellon University \quad  $^{2}$NEC Labs America \quad $^{3}$UC San Diego}

\begin{document}
\nolinenumbers   
\let\maketitle\maketitleold   

\maketitle

\begin{figure}[t]
    \centering
    \includegraphics[width=\textwidth]{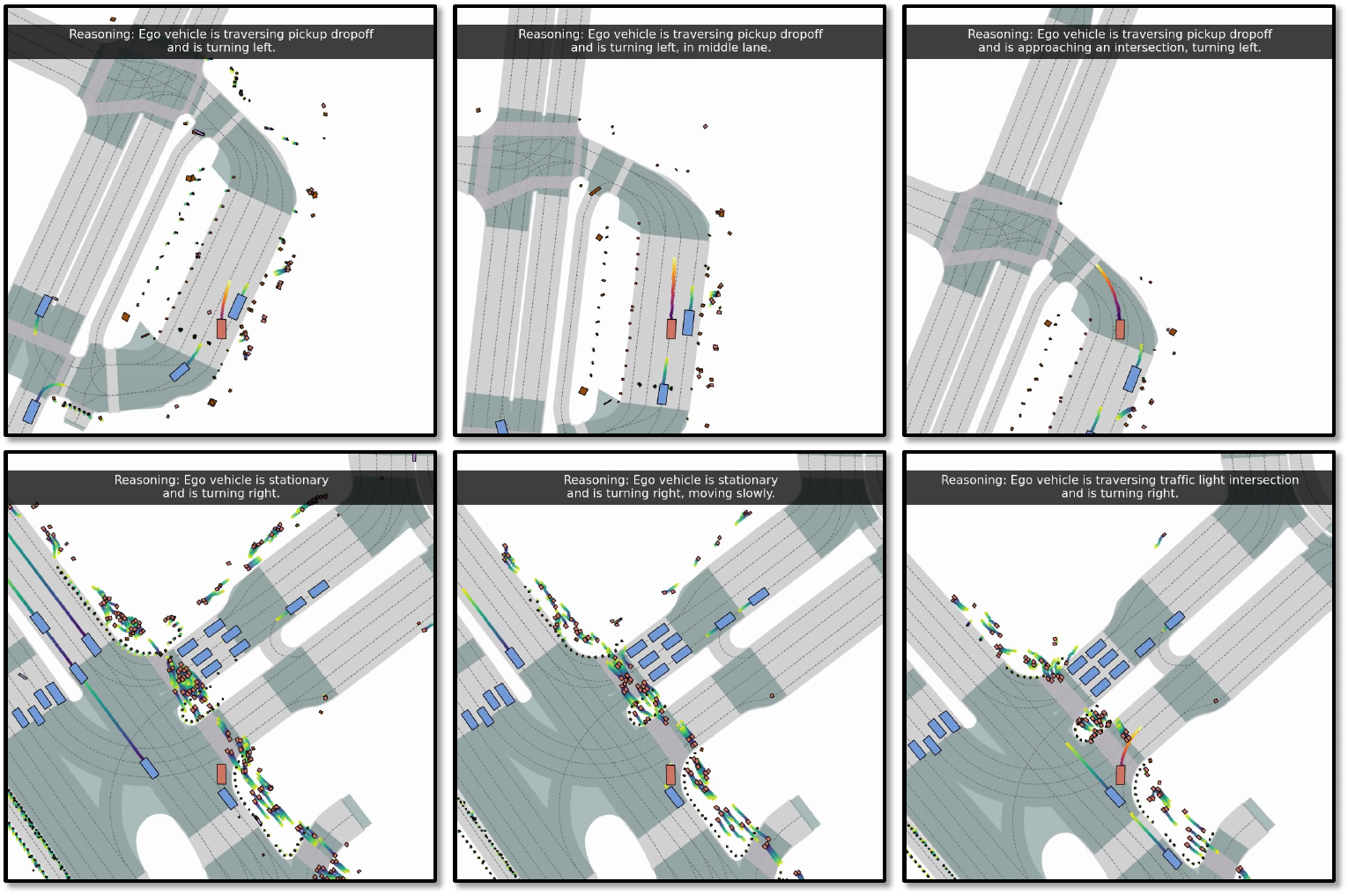}
    \caption{%
\textbf{Autonomous driving requires rule-following and semantic understanding.} \textit{(Top row)} A left turn at a pickup/dropoff zone: the ego vehicle (red, with planned trajectory) must navigate around vehicles blocking the lane.
\textit{(Bottom row)} A right turn through a crowded intersection: dense traffic from multiple directions, pedestrian crossings,
and ambiguous right-of-way require reasoning beyond simple trajectory optimization.
Text overlays show \lad's real-time situational understanding. The key insight is many scenarios labeled ``hard'' require only better lane-changing which our rule-based planner \rad can handle. Semantic difficulty demands language-grounded reasoning, e.g. negotiation of ambiguous traffic which \lad handles by generating both motion plans and interpretable explanations at $\sim$10Hz, enabling real-time deployment.%
}
    \label{fig:teaser}
    \vspace{-2em}
\end{figure}

\begin{abstract}
We present \lad, a real-time language--action planner with an interruptible architecture that produces a motion plan in a single forward pass ($\sim$20\,Hz) or generates textual reasoning alongside a motion plan ($\sim$10\,Hz). \lad is fast enough for real-time closed-loop deployment, achieving $\sim$3$\times$ lower latency than prior driving language models while setting a new learning-based state of the art on nuPlan Test14-Hard and InterPlan. We also introduce \rad, a rule-based planner designed to address structural limitations of PDM-Closed. \rad achieves state-of-the-art performance among rule-based planners on nuPlan Test14-Hard and InterPlan. Finally, we show that combining \rad and \lad enables hybrid planning that captures the strengths of both approaches. This hybrid system demonstrates that rules and learning provide complementary capabilities: rules support reliable maneuvering, while language enables adaptive and explainable decision-making.
\end{abstract}

\section{Introduction}
\label{sec:intro}

Consider the driving scenarios in \cref{fig:teaser}.
In the top row, an autonomous vehicle navigates a left turn at a pickup zone, a maneuver requiring awareness of lane blockages and intersection geometry. In the bottom row, the same vehicle attempts a right turn through dense, ambiguous traffic where pedestrians, oncoming vehicles, and unclear right-of-way create genuine uncertainty. Both scenarios appear in current closed-loop planning benchmarks, yet differ profoundly in character. The first demands safe maneuvering to navigate tight spaces; the second requires situational understanding to interpret intent and ambiguity.

Improvements to rule-based systems likely cannot address the true semantic long-tail, including situations requiring nuanced understanding of social norms, ambiguous occlusions, 
or negotiable rights-of-way~\cite{ghosh2025roadwork}.
For this, we introduce \lad (Language-Based Autonomous Driving),
a multimodal large language model (MLLM) planner built for real-time closed-loop deployment.
A persistent concern with language-based planners has been their latency~\cite{hwang2024emma, tian2024drivevlm, chen2024asyncdriver}: prior systems operate at 2-3\,Hz, far too slow for reactive closed-loop planning, leading some approaches to employ language models as offline advisors~\cite{sharan2023llm, chen2024asyncdriver}.
\lad's \emph{interruptible inference} architecture addresses this by producing a valid plan in a single forward pass ($\sim$20Hz) and optionally generating textual reasoning when compute budget permits ($\sim$10Hz), remaining compatible with safety mechanisms that require immediate re-planning (Section~\ref{sec:interruptible}).
Beyond real-time feasibility, our ablations show that language supervision provides complementary training signal for closed-loop planning, yielding strong performance on long-tailed benchmarks including nuPlan Test14-Hard and InterPlan.

We also design \rad (Rule-Based Autonomous Driving), a structured planner that extends PDM-Closed~\cite{dauner2023pdmclosed} with dynamic topology replanning and goal-directed optimization. \rad reveals that many scenarios labeled ``hard'' in current benchmarks are geometric in character, resolvable with capabilities like lane changes, while truly difficult cases require the semantic reasoning that language enables. Our hybrid planner, \rad-\lad, combines strict rule-following with language-based reasoning for better performance in long-tailed scenarios.

Thus, we view autonomous driving planning as consisting of two complementary challenges: geometric feasibility and semantic reasoning. RAD addresses the geometric component by expanding the planner's search space through dynamic topology and maneuver priors. LAD addresses the semantic component by enabling language-grounded reasoning over ambiguous traffic interactions. The resulting hybrid system provides a practical pathway for integrating structured planning and foundation models in real-time autonomy.

\noindent
In summary, our key contributions are:
\begin{enumerate}
\item \textbf{\lad}: The first real-time language-action planner to achieve state-of-the-art performance on closed-loop, long-tailed autonomous driving benchmarks. \lad demonstrates not only that language-based supervision improves planning performance, but also that inference-time, language-based reasoning is deployable at $\sim$10Hz.
\item \textbf{\rad}: A flexible rule-based planner that achieves strong performance on long-tail benchmarks by extending the capabilities of existing rule-based planners such as PDM-Closed~\cite{dauner2023pdmclosed}, revealing that much of the benchmark difficulty is geometric rather than semantic.
\item \textbf{\rad-\lad}: An integrated hybrid rule-and-language-based planner that combines the best of both worlds -- the interpretable language-based reasoning and planning of \lad and the physics-based safety guardrails of \rad -- to achieve competitive closed-loop planning performance.
\end{enumerate}

\section{Related Work}
\label{sec:related}

\subsection{Language-Based Planning}

Language-based planning offers a potential solution for the semantic reasoning gap left by rule and conventional learning-based planners
Approaches like DriveVLM~\cite{tian2024drivevlm}, DriveGPT4~\cite{xu2024drivegpt4}, and EMMA~\cite{hwang2024emma} have demonstrated strong scene understanding and reasoning capabilities. 
However, these systems are fundamentally limited by latency and open-loop design. Most operate at speeds (e.g., 2--3Hz) insufficient for reactive closed-loop planning or rely on offline processing~\cite{tiantokenize, wu2024smart}. 
Advisory frameworks~\cite{sharan2023llm, chen2024asyncdriver} attempt to mitigate this by decoupling reasoning from planning, but this prevents true language-guided improvisation. 

A common limitation across these methods is that plan generation is tightly coupled to full autoregressive text generation, making latency proportional to reasoning depth. \lad addresses this with an \emph{interruptible inference} architecture: a dedicated plan token always produces a valid trajectory in a single forward pass, while optional reasoning tokens precede it to improve planning quality when compute budget permits. A phased training curriculum (inspired by BLIP-2~\cite{li2023blip}, LLaVA~\cite{liu2023visual}, Pi~\cite{black2025pi0}) enables 10Hz planning with reasoning, solving the latency bottleneck that hindered the deployment of prior vision-language-action models.

\subsection{Rule-Based and Learned Planning}

Evaluation in autonomous driving has shifted from open-loop metrics to closed-loop simulation~\cite{caesar2021nuplan}, revealing that many state-of-the-art learning-based planners struggle to match the reliability of rule-based systems like PDM-Closed~\cite{dauner2023pdmclosed}.
While PDM-Closed excels on standard benchmarks, its fixed topology means that ``long-tail'' benchmarks~\cite{cheng2024plantf,hallgarten2024can} partially reflect its design constraints.
\rad addresses these with dynamic replanning and goal-directed optimization (Section~\ref{sec:method}), significantly outperforming both PDM-Closed and recent hybrid extensions~\cite{sun2024generalizingmotionplannersmixture, huang2023gameformer}.

\begin{wrapfigure}{r}{0.4\linewidth}
    \centering
    \includegraphics[width=\linewidth]{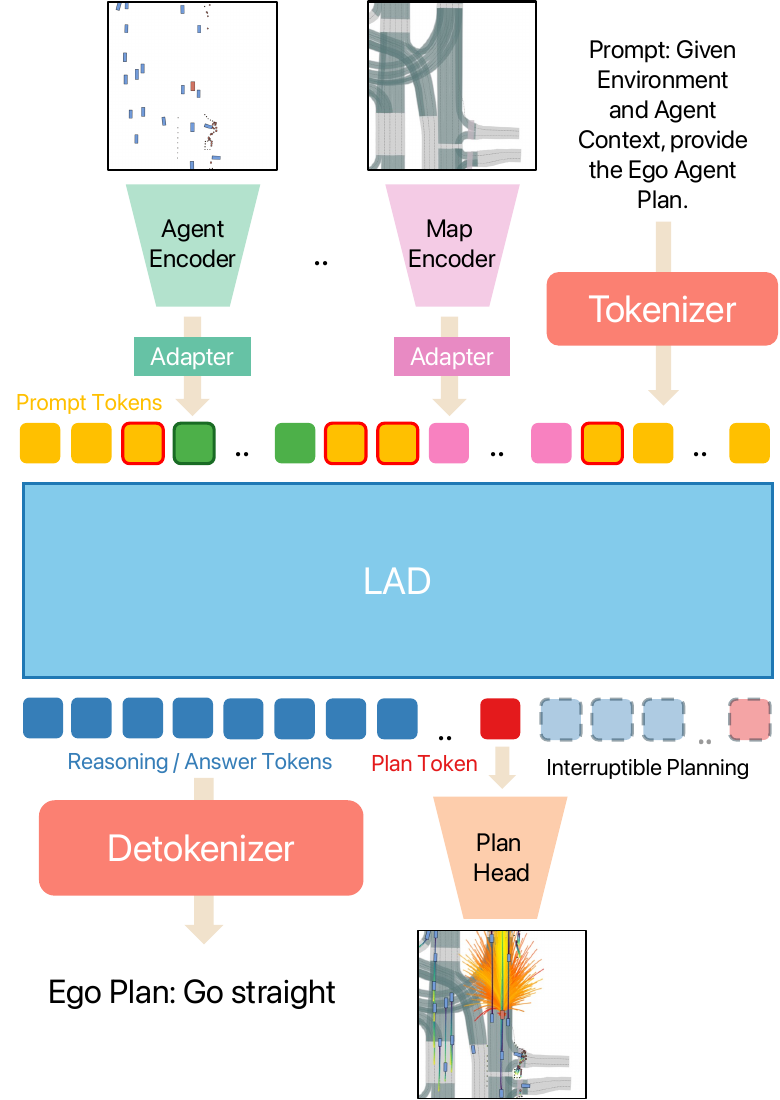}
    \caption{\small \textbf{\lad Architecture.} We encode scene context, adapt it into the language model's manifold, and insert it as pseudo-tokens within the prompt. The decoder produces natural-language reasoning and a motion plan from the hidden state at \tok{plan}.}
    \label{fig:planner-arch}
    \vspace{-4em}
\end{wrapfigure}

Pure imitation-based learned planners (e.g., PlanTF~\cite{cheng2024plantf}, DiffusionPlanner~\cite{zheng2025diffusionplanner}) offer promise for generalization but often exhibit poor adherence to safety constraints in closed-loop settings. Reinforcement learning offers a complementary direction: CaRL~\cite{jaeger2025carl} trains an action-based RL policy that bypasses the trajectory-to-control interface entirely. As output representation and training paradigm are tightly coupled in this setting, we treat this as an orthogonal axis and focus on improving planning through language supervision (see Supplemental Material for discussion).

\subsection{Hybrid Planning}

The complementary strengths of rule-based and learned planners have motivated hybrid approaches that combine both.
PLUTO~\cite{cheng2024pluto} augments a learned trajectory predictor with a rules-based scorer, and STR2~\cite{sun2024generalizingmotionplannersmixture} extends this with larger-scale mixture-of-experts architectures.
Similarly, DiffusionPlanner~\cite{zheng2025diffusionplanner} and FlowPlanner~\cite{tan2025flow} generate trajectories via generative models and refine them with rule-based scoring.
\rad-\lad follows this hybrid paradigm but integrates a learned language-based planner, enabling interpretable reasoning alongside rule compliance.

\section{Method}
\label{sec:method}

\subsection{\lad: Language-Based Autonomous Driving}

\lad is an anytime multimodal language model planner - 
it produces a valid motion plan in a single forward pass
and optionally generates textual reasoning when computational budget permits,
enabling real-time closed-loop deployment. Here, ``multimodal'' refers to the fusion of structured scene entities (vectorized map and agent representations) with language.

\subsubsection{Architecture.}

We transform a pretrained decoder-only language model into a motion planner
by introducing three modifications:
(1) scene encodings are injected as pseudo-tokens,
(2) a planning head is attached to a designated output position,
and (3) inference can be interrupted to meet latency constraints.

\paragraph{Scene encoding and adaptation.}
Let $\mathcal{M}=\{m_i\}_{i=1}^{N_m}$ denote the set of map elements (lanes, crosswalks),
and $\mathcal{A}=\{a_j\}_{j=1}^{N_a}$ denote dynamic and static agents.
Both are encoded via PlanTF~\cite{cheng2024plantf},
\begin{align}
\mathbf{z}_i &= \phi_{\text{PlanTF}}^{\text{map}}(m_i) \in \mathbb{R}^{d_\text{ptf}}, \quad i=1,\dots,N_m, \\
\mathbf{u}_j &= \phi_{\text{PlanTF}}^{\text{agent}}(a_j) \in \mathbb{R}^{d_\text{ptf}}, \quad j=1,\dots,N_a.
\end{align}
Lightweight MLP adapters project these embeddings into the model's token space:
\begin{align}
\tilde{\mathbf{z}}_i = f_{\text{map}}(\mathbf{z}_i) \in \mathbb{R}^{d_\ell}, \qquad
\tilde{\mathbf{u}}_j = f_{\text{agent}}(\mathbf{u}_j) \in \mathbb{R}^{d_\ell}.
\end{align}
While we instantiate \lad with PlanTF for structured inputs,
the same adapter pattern extends to vision encoders for camera or lidar modalities.

\paragraph{Multimodal prompting.}
Following object-centric tokenization~\cite{tiantokenize},
we inject adapted embeddings as pseudo-tokens delimited by special tokens
and interleave them with a natural-language task prompt.
The decoder-only language model attends over this heterogeneous context as it would attend over a purely textual input.

\paragraph{Planning Decoder.}
Rather than generating waypoints autoregressively,
we formulate planning as classification over a discrete trajectory vocabulary following prior works~\cite{shi2022motion, li2024hydramdp,wu2024smart},
$\mathcal{V}=\{\mathbf{v}_k\}_{k=1}^{K}$, where $K$ is the vocabulary size and each prototype $\mathbf{v}_k\in\mathbb{R}^{T\times 2}$ is a trajectory of $T$ waypoints.
A small MLP head $g:\mathbb{R}^{d_\ell}\!\rightarrow\mathbb{R}^{K}$
at the \tok{plan} token produces logits $\mathbf{s}=g(\mathbf{h}_{\text{plan}})$. The classifier is trained with the imitation soft cross-entropy loss~\cite{li2024hydramdp}, where targets are derived from the proximity of each prototype to the ground-truth trajectory $\mathbf{v}^{*}$,
\begin{align}
y_k=\frac{\exp\!\bigl(-\lVert\mathbf{v}_k-\mathbf{v}^{*}\rVert^2\bigr)}{\sum_{j=1}^{K}\exp\!\bigl(-\lVert\mathbf{v}_j-\mathbf{v}^{*}\rVert^2\bigr)},\qquad
\mathcal{L}_{\text{plan}}=\log\!\sum_{k=1}^{K}e^{s_k}\;-\;\sum_{k=1}^{K}y_k\,s_k.
\end{align}

The classification head follows BERT or ViT class-token heads~\cite{devlin2019bert,dosovitskiy2021vit} and GPT's task-specific output heads~\cite{radford2018improving}, adapted to a decoder-only context.

\paragraph{Textual Supervision.}

When reasoning (or answer) text is available, we train the model with teacher forcing over response tokens,
\begin{align}
\mathcal{L}_{\text{language}}
\;=\;
-\sum_{t\in\Omega}\log p_\theta\!\big(w^\star_t \,\big|\, w^\star_{<t},\, \mathbf{X}\big),
\end{align}
where, $\Omega$ indexes response tokens (excluding prompt and special tokens)
and $\mathbf{X}$ is the multimodal input sequence.
The planning head is trained jointly via cross-entropy over the trajectory vocabulary.

\paragraph{Overall Objective.}
The full training loss is a sum of the above terms,
\begin{align}
\mathcal{L}=\mathcal{L}_{\text{plan}} + \mathcal{L}_{\text{language}},
\end{align}
and we observed hand-engineered weighting schemes hurt performance.

\subsubsection{Interruptible Anytime Inference.}
\label{sec:interruptible}

Real-time deployment imposes hard latency deadlines that may vary with speed, traffic density, and safety interventions. A planner that couples action generation to lengthy reasoning chains cannot meet these constraints reliably. Our interruptible inference architecture addresses this: a valid motion plan is always available from the plan token, while reasoning tokens are generated opportunistically when budget permits.

\begin{wrapfigure}{r}{0.5\linewidth}
    \centering
    \vspace{-1em}
    \includegraphics[width=0.8\linewidth]{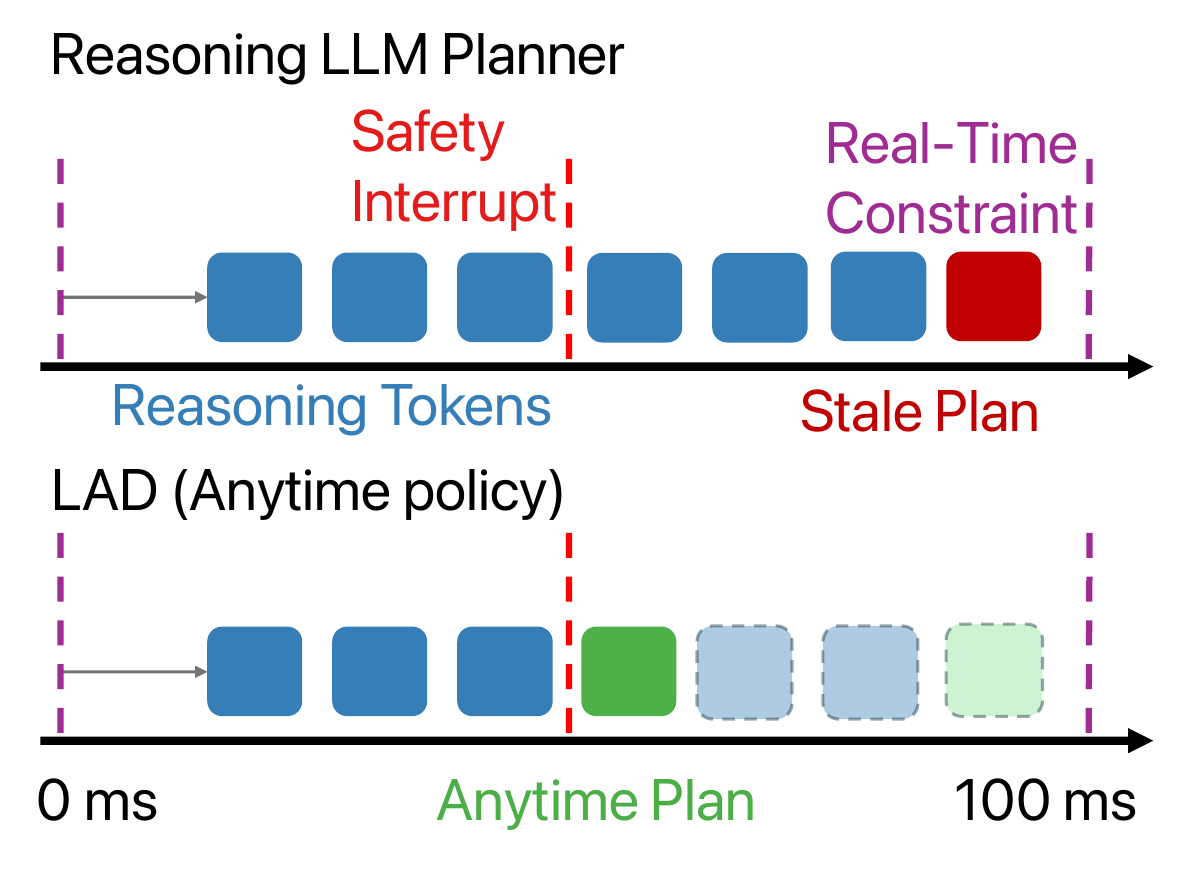}
    \caption{\small \textbf{Interruptible Anytime Inference.} \lad produces a valid plan when interrupted, generating reasoning tokens if budget permits.}
    \label{fig:anytime-planning}
    \vspace{-2em}
\end{wrapfigure}

The key observation is that reasoning text is optional for trajectory prediction. When latency constraints are strict,
we append \tok{plan} directly after the scene tokens
and perform a single prefill pass to obtain $\mathbf{h}_{\text{plan}}$,
yielding a motion plan with one forward pass.
When computational budget permits,
we allow the model to generate reasoning tokens before inserting \tok{plan},
trading latency for interpretability.
The architecture remains identical in both cases; only the prompt structure changes.
More broadly, because the plan token yields a valid action whether or not reasoning tokens precede it, this architecture is compatible with deployment where a safety mechanism requires an immediate action.

This design enables \lad to operate in real-time without reasoning or with short justifications,
while remaining compatible with standard inference optimizations
including KV-caching~\cite{pope2023efficiently},
operator fusion~\cite{Spector2025_NoBubbles},
and efficient scheduling~\cite{kwon2023vllm}.
Detailed latency analysis is provided in Section~\ref{sec:rt_results}.

\subsubsection{Multimodal Training}
\label{sec:multi-train}

Training a multimodal planner requires balancing
the preservation of language representations
while acquiring new capabilities. We adopt the following training curriculum to achieve this balance.

\paragraph{Stage A: Alignment.}
The language model remains frozen while only lightweight adapters and multimodal projection layers are trained. This establishes stable grounding of scene encodings within the model's existing representational space, 
following alignment strategies in prior multimodal work~\cite{li2023blip,liu2023visual}.

\paragraph{Stage B: LoRA finetuning.}
LoRA modules~\cite{hu2022lora} and the task-specific planning head are introduced while the backbone remains frozen. The zero-initialized updates of LoRA provide controlled capacity expansion for semantic grounding and trajectory prediction without destabilizing pretrained representations.

\noindent
To maintain linguistic abilities of \lad, we employ a mixed training strategy that co-trains on a small proportion of the Interaction QA dataset alongside the planning objective, following prior work~\cite{zitkovich2023rt,black2025pi0} and mitigating catastrophic forgetting of pretrained representations~\cite{que2024d}.

\subsection{\rad: Rule-Based Autonomous Driving}

State-of-the-art rule-based planners like PDM-Closed~\cite{dauner2023pdmclosed}
perform remarkably well on standard scenarios (i.e., Val14~\cite{caesar2021nuplan}) but score badly on long-tailed scenarios (Test14-Hard~\cite{cheng2024plantf} and InterPlan~\cite{hallgarten2024can}).
These failures stem from specific design choices in the baseline (e.g., fixed topology, no deadlock handling) rather than intrinsic scenario complexity.
\rad addresses PDM-Closed's key limitations such as static topology and no lane changes with
\emph{dynamic topology replanning},
\emph{lane-change capability}, and \emph{goal-directed optimization}.

\subsubsection{Revisiting Rule-Based Planning in PDM.}

PDM-Closed~\cite{dauner2023pdmclosed} planner selects the optimal trajectory $\pi^*$ by maximizing a scoring function. To align with the official nuPlan evaluation metrics, this is formulated as a cost function $J_{\text{PDM}}$ composed of multiplicative penalties (safety constraints) scaling a weighted sum of driving quality objectives,

\begin{equation}
    J_{\text{PDM}}(\pi) = C_{\text{col}} C_{\text{ra}} C_{\text{mp}} \bigg(
     w_{\text{ttc}} C_{\text{ttc}} + w_{\text{dr}} C_{\text{dr}} + w_{\text{sp}} C_{\text{sp}} + w_{\text{ep}} C_{\text{ep}} + w_{\text{cf}} C_{\text{cf}} \bigg)
\label{eq:pdm_cost}
\end{equation}

Terms of $J_{\text{PDM}}$ include multiplicative penalties for collision $C_{\text{col}}$, violating drivable area $C_{\text{ra}}$, not making minimum progress $C_{\text{mp}}$, and weighted costs for time-to-collision $C_{\text{ttc}}$, speed compliance $C_{\text{sp}}$, rogress along the experts’ route $C_{\text{ep}}$, direction compliance $C_{\text{dr}}$, and comfort $C_{\text{cf}}$.

While robust on standard scenarios, PDM-Closed~\cite{dauner2023pdmclosed} has a fixed topology: (1) it does not support lane changes; (2) it enforces strict penalties hindering necessary evasive maneuvers; and (3) it does topological planning only once at the start, causing drift. \rad addresses these limitations via the following modifications.

\subsubsection{Dynamic Topological Replanning.}

PDM-Closed~\cite{dauner2023pdmclosed} generates 15 candidates per timestep but anchors them to proposal paths ($\Gamma_{\text{static}}$) \textit{fixed at initialization}, i.e. there is no topological replanning after initialization. If the ego deviates or paths become blocked, proposals are never updated.

\rad performs \emph{full topological replanning at every timestep}. We define the proposal paths extraction as a time-dependent function of the current ego state $\mathbf{s}_t$ and the map $\mathcal{M}$,
\begin{equation}
    \Gamma_t = \text{GraphSearch}(\mathbf{s}_t, \mathcal{M}).
\end{equation}

\begin{wrapfigure}{r}{0.5\linewidth}
    \centering
    \includegraphics[width=\linewidth]{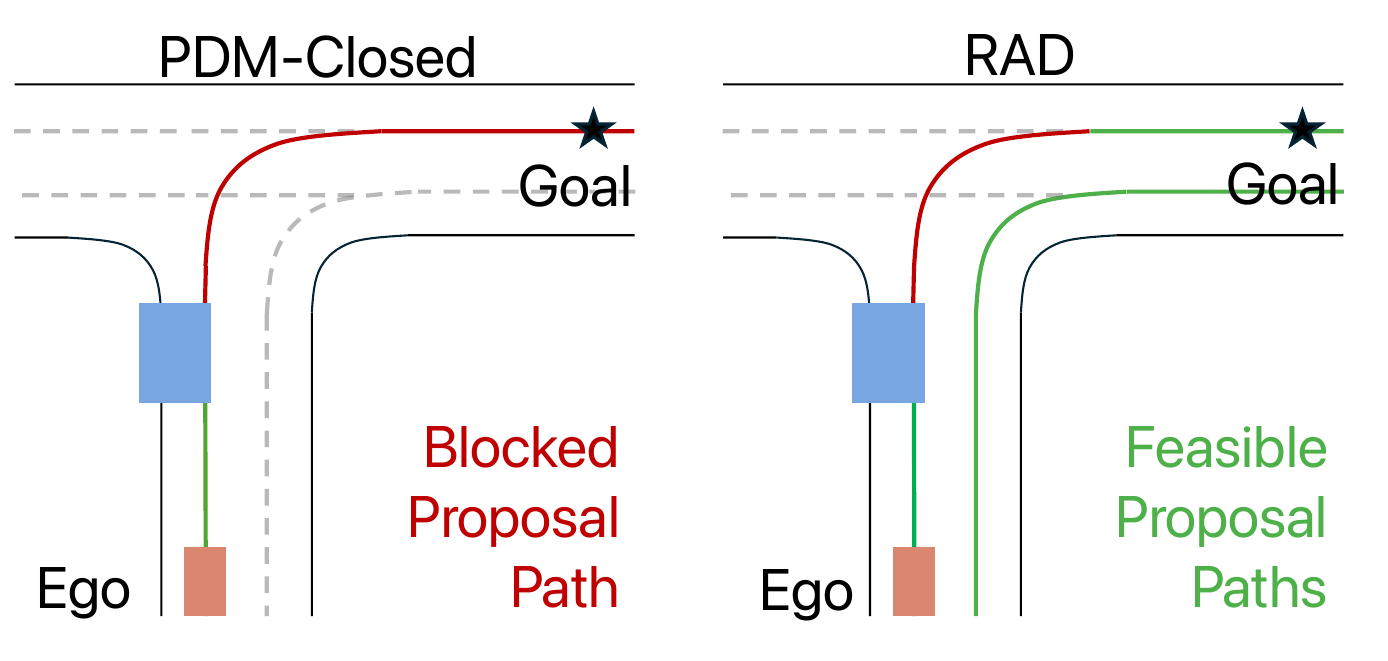}
    \vspace{-2em}
    \caption{\small PDM-Closed's~\cite{dauner2023pdmclosed} static proposal paths become blocked by obstacles with no recovery. \rad topologically replans at every timestep and augments the route with adjacent-lane centerlines.}
    \label{fig:proposal-paths}
    \vspace{-3em}
\end{wrapfigure}

Consequently, the set of available trajectory proposals $\Pi_t$ is dynamically updated to reflect the instantaneous topology,
\begin{equation}
    \Pi_t = \bigcup_{\gamma \in \Gamma_t} \bigcup_{o \in \mathcal{O}} \text{IDM}(\mathbf{s}_t, \gamma, o, v_0).
\end{equation}

Here, $\mathcal{O}$ denotes discrete lateral offsets and $v_0$ is the IDM reference velocity.
This ensures the optimization horizon always extends from the vehicle's actual current pose, allowing for robust recovery if the vehicle is forced off the nominal path.

\subsubsection{Lane-Changing via Topology Augmentation.}

To enable lane changes, \rad augments the road topology with \emph{adjacent-lane centerlines}. While PDM-Closed~\cite{dauner2023pdmclosed} considers a single route-based centerline $\Gamma_{\text{ego}}$, \rad expands this to include spatially adjacent centerlines $\Gamma_{\text{adj}}$, even those with opposing traffic flow,
\begin{equation}
    \Gamma_{\text{RAD}} = \{ \Gamma_{\text{ego}} \} \cup \{ \Gamma_{\text{left}}, \Gamma_{\text{right}} \} \cup \Gamma_{\text{opp}}.
\end{equation}
The proposal set is then expanded to sample trajectories relative to all centerlines in this augmented set,
\begin{equation}
    \Pi_{\text{aug}} = \bigcup_{\gamma \in \Gamma_{\text{RAD}}} \bigcup_{o \in \mathcal{O}} \text{IDM}(\mathbf{s}_t, \gamma, o, v_0),
\end{equation}

where $\mathcal{O}$ denotes the set of discrete lateral offsets. This allows the planner to sample from a richer family of trajectories, including feasible lane-change proposals.

\subsubsection{Goal-Directed Optimization.}

\rad modifies the objective to encourage decisive progress toward the mission goal. Instead of relying solely on path-integrated distance, \rad computes a \emph{Euclidean distance-to-goal} cost. Let $\mathbf{p}_T^\pi$ be the position of the ego vehicle at the end of planning horizon $T$ for proposal $\pi$, and $\mathbf{g}$ be the global goal coordinates,
\begin{equation}
    J_{\text{goal}}(\pi) = \| \mathbf{p}_T^\pi - \mathbf{g} \|_2.
\end{equation}
The total cost function $J_{\text{RAD}}$ linearly combines the baseline PDM cost with this goal-seeking term,
\begin{equation}
    J_{\text{RAD}}(\pi) = J_{\text{PDM}}(\pi) + w_{\text{goal}} J_{\text{goal}}(\pi).
\end{equation}
This optimization encourages advancement in open regions and helps escape local minima induced by complex road geometries.

\subsubsection{Trajectory Proposal Augmentation via Vocabulary.}

To diversify candidate trajectories beyond geometric centerlines, \rad incorporates all $K$ proposals from a precomputed \emph{trajectory vocabulary}. We construct a vocabulary $\mathcal{V}=\{\mathbf{v}_k\}_{k=1}^{K}$ by clustering ego trajectories from the nuPlan training set, following prior works~\cite{shi2022motion, li2024hydramdp,wu2024smart}. The final proposal set $\Pi_{\text{RAD}}$ is the union of topologically augmented IDM~\cite{treiber2000congested} proposals and the data-driven vocabulary,
\begin{equation}
    \Pi_{\text{RAD}} = \Pi_{\text{aug}} \cup \{ T_{\text{ego}}(\mathbf{v}) \mid \mathbf{v} \in \mathcal{V} \}.
\end{equation}
This injects \emph{data-driven maneuver priors} (e.g., swerves, bypasses) into the rule-based system.

\subsubsection{Context-Aware Rule Relaxation.}

\begin{wrapfigure}{r}{0.5\linewidth}
    \centering
    \vspace{-3em}
    \includegraphics[width=\linewidth]{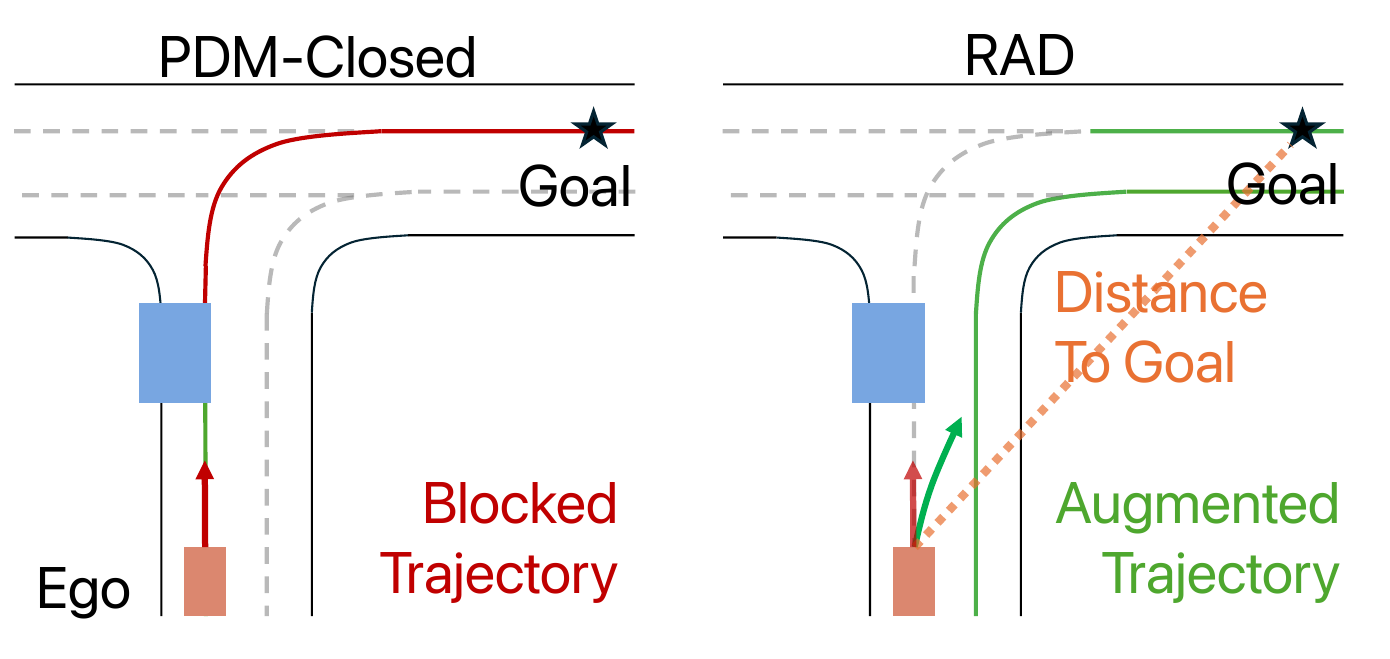}
    \vspace{-2em}
    \caption{\small \rad combines goal-directed optimization with trajectory proposal augmentation, adding feasible alternative trajectories and favoring trajectories that make progress toward the goal.}
    \label{fig:goal-directed-vocab}
    \vspace{-2em}
\end{wrapfigure}

To handle deadlock situations, \rad introduces a \emph{soft rule relaxation} mechanism. We define a relaxation indicator $\mathbb{I}_{\text{relax}} \in \{0, 1\}$, active when blockage is detected. When active, \rad contextually downweights penalties for driving-direction and driving-area violations, allowing the optimizer to consider safe, short-term deviations from nominal traffic rules to circumvent obstacles.

\subsection{Hybrid Planner Integration}

While \lad and \rad are state-of-the-art methods in their respective class of methods, that is, within learned and rule-based planners, creating a hybrid planner requires careful co-design accounting for the downstream LQR controller and rules scorer. 

\noindent
\textbf{Trajectory Refinement Head.} Adding a trajectory refinement module provides \lad the ability to refine its trajectory for better rules alignment for hybrid integration. The classifier's argmax selects a coarse prototype $\hat{\mathbf{v}}\in\mathbb{R}^{T\times 2}$, which is concatenated with the plan token embedding and passed through an MLP head $r:\mathbb{R}^{d_\ell+2T}\rightarrow\mathbb{R}^{2T}$ to predict per-waypoint offsets, $$\tilde{\mathbf{v}}=\hat{\mathbf{v}}+r\bigl([\mathbf{h}_{\text{plan}}\oplus\operatorname{vec}(\hat{\mathbf{v}})]\bigr)$$ 

Crucially, the final layer of $r$ is zero-initialised~\cite{zhang2023adding} so that the trajectory refinement head acts as the identity at the start of training and does not interfere with the classifier's learning signal. The refinement loss is, 
\begin{align}
\mathcal{L}_{\text{refine}}=\frac{1}{T}\sum_{t=1}^{T}\lVert\tilde{\mathbf{v}}_t-\mathbf{v}_t^{*}\rVert_2.
\end{align}

\noindent
\textbf{Rules-Based Refinement.} Following prior work~\cite{huang2023gameformer,
sun2024generalizingmotionplannersmixture, zheng2025diffusionplanner} we adopt a cost-based refinement approach, expanding the vocabulary of the rule-based planner by adding learning based planner's plan as an additional proposal, further applying offsets to the model outputs to augment the rule-based planner with additional candidate trajectories. All the candidate trajectories are then scored using a rules scorer, which can be either PDM-Closed's~\cite{dauner2023pdmclosed} or RAD's scorer.

\section{Experimental Details}
\label{sec:setup}

\paragraph{Datasets.} We train and evaluate our methods on the \textbf{nuPlan} simulator and dataset~\cite{caesar2021nuplan} which is a closed-loop simulator grounded in real-world driving logs. We do not focus on NavSim~\cite{dauner2024navsim} as it does not perform true closed-loop evaluation and CARLA~\cite{dosovitskiy2017carla} which lacks realistic logged driving data. We train on 1 million nuPlan scenarios following parity with prior work~\cite{cheng2024plantf, cheng2024pluto, zheng2025diffusionplanner, tan2025flow}. To provide diverse reasoning supervision, we construct \textbf{DrivingQA} and \textbf{PlanningQA} (See \supplref{sec:supp_datasets} for more details), synthesizing QA pairs from InterDrive~\cite{chang2025langtraj} behavior annotations using Qwen2.5-32B~\cite{qwen25vl} model and ground trajectories with textual reasoning. This synthetic dataset generates instructions grounded in valid trajectories, bridging the gap between raw behavioral data and semantic reasoning.

\begin{figure*}[t]
    \includegraphics[width=\textwidth]{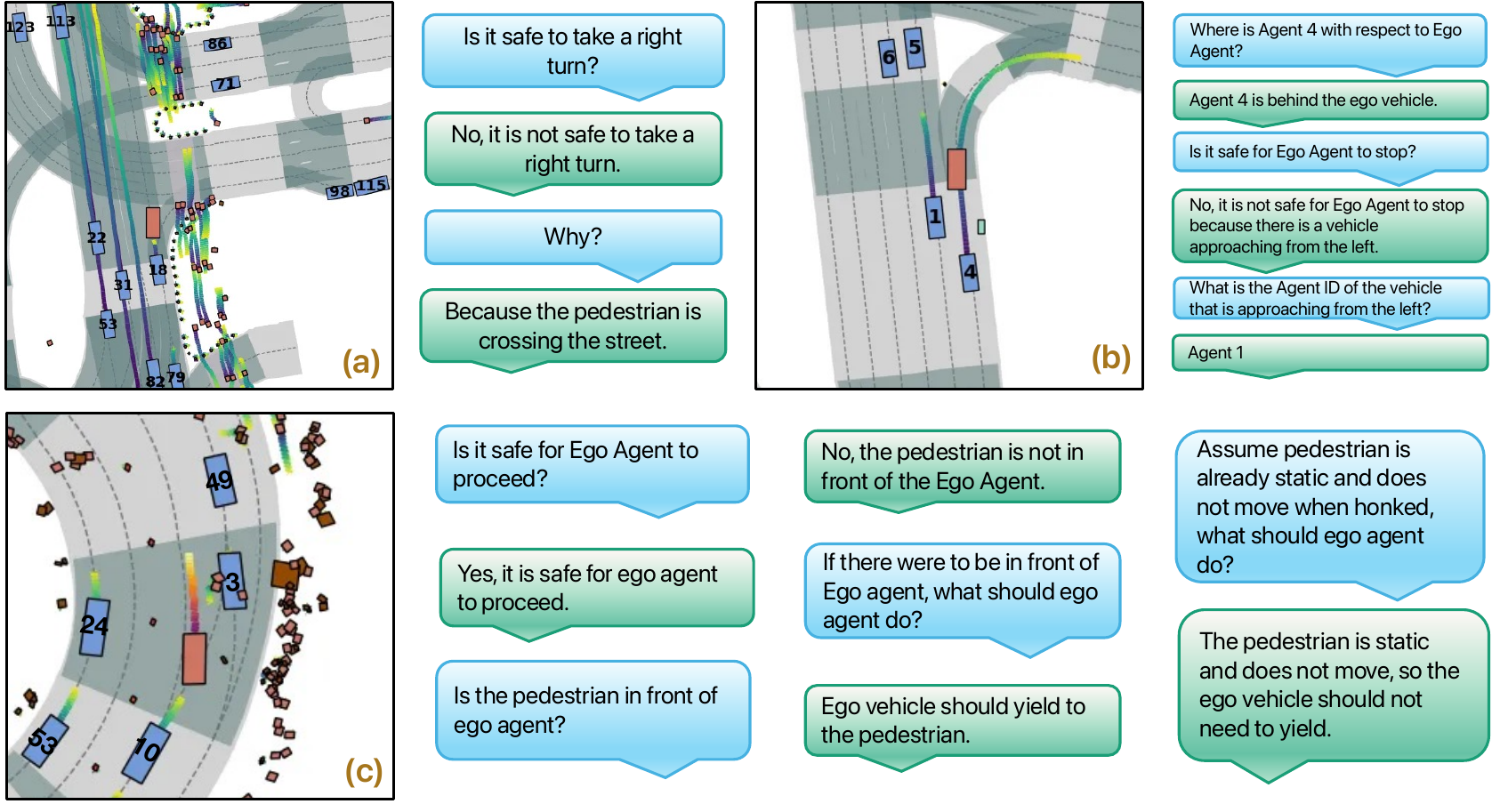}
    \caption{\small \textbf{Scenario Understanding and Reasoning.} (a) \textit{Cross-modal grounding}: \lad can interpret pedestrians, lane geometry, and turn-safety from scene context. (b) \textit{Relational reasoning}: \lad can infer spatial relations between agents and anticipates conflicts. (c) \textit{Conditional reasoning}: \lad can evaluate counterfactuals, apply traffic laws, and identify when rules may be safely relaxed.}
    \vspace{-2em}
    \label{fig:chatbot-examples}
\end{figure*}

\paragraph{Benchmarks.} We focus our evaluation on two challenging benchmarks:
\begin{itemize}
    \item \textbf{nuPlan Test14-Hard}~\cite{cheng2024plantf}, a subset of 14 scenario types specifically filtered to include difficult cases where the PDM-Closed baseline fails.
    \item \textbf{InterPlan}~\cite{hallgarten2024can}, a synthetic benchmark designed to test multi-agent interaction and deadlock resolution in driving.
\end{itemize}

\paragraph{Baselines.} We compare against three categories:
\begin{itemize}
    \setlength\itemsep{0em}
    \item \textbf{Rule-Based}: We evaluate our planners against PDM-Closed~\cite{dauner2023pdmclosed} (the nuPlan challenge winner) and IDM.
    \item \textbf{Learning-Based}: We compare against PlanTF~\cite{cheng2024plantf} (a pure Transformer approach), PLUTO~\cite{cheng2024pluto} (hybrid scoring), STR2~\cite{sun2024generalizingmotionplannersmixture} (a vision-centric raster-map-based approach), and Diffusion-Planner~\cite{zheng2025diffusionplanner} and FlowPlanner~\cite{tan2025flow}. These represent the current state-of-the-art in imitation learning based planners.
    \item \textbf{Multimodal Language-based Planners}: Unfortunately, no Closed-Loop Language-based Planners have been proposed for the nuPlan benchmark. Thus, we benchmark latency against DriveVLM~\cite{tian2024drivevlm} and DriveGPT4~\cite{xu2025drivegpt4} to contextualize \lad's real-time performance.
\end{itemize}

\paragraph{Implementation Details.}

\noindent
\textbf{\rad}: We implement all the changes to PDM-Closed such as topological replanning using the \textit{tuplan\_garage} framework.

\noindent
\textbf{\lad}: We use {Qwen3-0.6B}~\cite{yang2025qwen3} as the language backbone. To encode map, agents and decoding agent predictions, we employ PlanTF~\cite{cheng2024plantf}'s architecture.

\noindent
To achieve real-time performance with \lad, we implement a custom inference backend on a fork of \textit{nano-vllm}~\cite{GeeeekExplorer2025nanovllm} with KV-caching~\cite{pope2023efficiently} and operator fusion~\cite{Spector2025_NoBubbles}. All inference experiments are conducted on a single NVIDIA RTX A6000 to verify deployment feasibility.

\section{Experimental Results}
\label{sec:results}

We evaluate \rad, \lad, and their hybrid combination across nuPlan Test14 splits and the InterPlan benchmark.

\begin{table}[t]
\centering
\caption{\small \textbf{Test14-Hard (Reactive) and InterPlan results.} \rad rivals hybrid methods without learned components while \lad sets a new state-of-the-art among learned planners while offering textual reasoning ability.}
\vspace{-1em}
\small
\resizebox{0.6\columnwidth}{!}{%
\setlength{\tabcolsep}{5pt}
\begin{tabular}{llcc}
\toprule
\textbf{Type} & \textbf{Planner} & \textbf{Test14-Hard} & \textbf{InterPlan} \\
\midrule

\color{gray}{Expert}
& \color{gray}{Log Replay} & \color{gray}{85.96} & \color{gray}{--} \\

\midrule
\multirow{3}{*}{Rule}
& IDM~\cite{treiber2000congested} & 62.26 & 31 \\
& PDM-Closed~\cite{dauner2023pdmclosed} & 75.19 & 42 \\
& \textbf{\rad\ (Ours)} & 80.53 & 72 \\

\midrule
\multirow{5}{*}{Learned}
& PlanTF~\cite{cheng2024plantf} & 61.61 & 32 \\
& PLUTO~\cite{cheng2024pluto} & 59.74 & -- \\
& DiffusionPlanner~\cite{zheng2025diffusionplanner} & 69.22 & 25 \\
& FlowPlanner~\cite{tan2025flow} & 70.42 & -- \\
& \textbf{\lad\ (Ours)} & 70.77 & 40 \\

\midrule
\multirow{6}{*}{Hybrid}
& PLUTO~\cite{cheng2024pluto} & 76.88 & 49 \\
& STR2-CKS-800m~\cite{sun2024generalizingmotionplannersmixture} & 78.58 & 45 \\
& STR2-CPKS-800m~\cite{sun2024generalizingmotionplannersmixture} & 82.02 & 45 \\
& DiffusionPlanner~\cite{zheng2025diffusionplanner} & 82.00 & -- \\
& FlowPlanner~\cite{tan2025flow} & 80.25 & -- \\
& \textbf{RAD-\lad\ (Ours)} & 81.36 & 74 \\
\bottomrule
\end{tabular}
}
\label{tab:test14-hard}
\vspace{-2em}
\end{table}

\subsection{Performance on NuPlan Long-Tail Scenarios}
\label{sec:test14-results}

Table~\ref{tab:test14-hard} summarizes performance across Test14-Hard, and InterPlan. We first discuss Test14-Hard results.
\rad achieves 80.53 in the reactive setting, outperforming PDM-Closed by 5.34 points while rivaling hybrid methods. This gap arises not from sophisticated reasoning but from \rad's ability to perform lane changes and replan dynamically. PDM-Closed does not include these capabilities. Critically, as shown in Table~\ref{tab:val14} of the supplementary material, \rad also outperforms PDM-Closed on the standard nuPlan split (Val14), demonstrating that \rad is not overfit to long-tail scenarios.

This finding has important implications for how we interpret benchmark difficulty. The Test14-Hard split filters for PDM-Closed failures~\cite{cheng2024plantf}, implying that many ``hard'' cases can be resolved once a planner can change lanes or recover from off-route situations. \rad nearly matches the strongest prior hybrid methods without any learned components. 

Among learning-based approaches, \lad achieves the strongest reactive performance, surpassing all existing methods~\cite{cheng2024plantf, zheng2025diffusionplanner, tan2025flow}. 

\subsection{Generalization to Synthetic Long-Tail Scenarios}
\label{sec:interplan-results}

To evaluate generalization to more diverse long-tail conditions, we turn to InterPlan~\cite{hallgarten2024can}, a benchmark specifically designed to stress-test planners on more realistic long-tail scenarios.
InterPlan augments nuPlan scenarios with additional agents, obstacles, and alternative navigation goals, creating situations that require multi-agent coordination and deadlock resolution.

The InterPlan column of Table~\ref{tab:test14-hard} reveals several insights.
First, \rad dramatically outperforms all rule-based planners nearly doubling prior rule-based state of the art~\cite{dauner2023pdmclosed}.
This confirms that the architectural improvements in \rad (lane changes, goal-directed optimization, rule relaxation) provide broad benefits for long-tailed situations.

Second, \lad improves performance when compared to other learned closed-loop planners. This demonstrates that real-time language-based reasoning can meaningfully improve planning in interactive scenarios.

Our hybrid planner, \rad-\lad reaches 74, outperforming several other hybrid approaches. The complementary nature of rules and learning based planning is evident: rules handle robust maneuvering and deadlock resolution, while \lad contributes contextual reasoning for situations where rules alone are insufficient.

\subsection{Ablations}

Table~\ref{tab:lad-ablations} ablates the incremental contribution of each \lad component on Test14-Hard (Reactive), starting from the PlanTF~\cite{cheng2024plantf} baseline. For \rad ablations, please see \supplref{sec:supp_rad}.

The original PlanTF baseline considers only 32 dynamic agents; increasing this to 128 does not improve performance, likely because the additional distant agents introduce noise without providing useful planning signal. Adding static objects (e.g., barriers, cones) confirms that static scene context is critical for safe maneuvering, making PlanTF~\cite{cheng2024plantf} competitive with DiffusionPlanner~\cite{zheng2025diffusionplanner}, PLUTO~\cite{cheng2024pluto}, and FlowPlanner~\cite{tan2025flow}. This is consistent with recent findings that, for planning and control, architecture and input quality~\cite{cheng2024plantf} dominate over the choice of training objective~\cite{pan2025much}.

Introducing the plan token (Section~\ref{sec:method}) validates that single-step classification is effective for extracting waypoints from hidden state. Incorporating DrivingQA with our multimodal large language model, further improves performance by 0.85 points, indicating that diverse language supervision provides useful inductive bias for trajectory prediction even when the QA content is not directly conditioned on the planning output. Finally, adding PlanningQA with ego behavior text (scenario type, meta-action) that is temporally aligned with the ground-truth trajectory yields the final \lad model at 70.77, demonstrating that action-aligned textual supervision provides complementary learning signal for trajectory prediction.

\begin{table}[t]
\centering
\caption{\small \textbf{\lad component ablations on Test14-Hard (Reactive).} Each component contributes incrementally, with language supervision through DrivingQA and PlanningQA providing complementary gains on top of input and architectural improvements. Textual reasoning acts as a useful inductive bias for trajectory prediction.}
\vspace{-1em}
\small
\resizebox{0.4\columnwidth}{!}{%
\begin{tabular}{lc}
\toprule
\textbf{Component} & \textbf{Test14-Hard (R)} \\
\midrule
PlanTF & 61.61 \\
+ 128 Objects & 59.73 \\
+ Static Objects & 68.49 \\
+ Plan Token & 68.90 \\
+ LLM/DrivingQA & 69.75 \\
+ PlanningQA (\lad) & \textbf{70.77} \\
\bottomrule
\end{tabular}
}
\label{tab:lad-ablations}
\vspace{-2em}
\end{table}

\subsection{Real-Time Performance}
\label{sec:rt_results}

A persistent concern with closed-loop language-based planners is latency - prior literature have widely regarded these planners as too slow for closed-loop deployment~\cite{hwang2024emma, tian2024drivevlm, chen2024asyncdriver, jiang2025survey}. Some prior work sidestep this failure-mode entirely by using language models only as offline advisors~\cite{sharan2023llm, chen2024asyncdriver}. We show this trade-off may not be necessary and Table~\ref{tab:latency} shows that this limitation is not fundamental.

In contrast, \lad operates at 20Hz (43ms) without reasoning and maintains approximately 10Hz operation (102ms) with 10 output tokens which is sufficient for real-time justifications. Note that without reasoning \lad runs faster than DiffusionPlanner~\cite{zheng2025diffusionplanner} and FlowPlanner~\cite{tan2025flow}, which produce no textual output at all. Our current implementation leaves significant room for further acceleration through orthogonal optimizations such as improved quantization strategies~\cite{zandieh2025turboquant}, suggesting that the gains reported here represent a lower bound.

\subsection{Qualitative Analysis}
\label{sec:qualitative}

Beyond quantitative metrics, \lad's multimodal phased training (Section~\ref{sec:multi-train}) enables it to reason conversationally about driving scenarios.
Figure~\ref{fig:chatbot-examples} illustrates three capabilities essential for a language-based planner.

\begin{table}[t]
\centering
\caption{\small \textbf{\lad achieves real-time language-based planning.} Without reasoning, \lad runs at 43\,ms ($\sim$20\,Hz), comparable to recent closed-loop planners. With reasoning enabled (10 tokens), \lad operates at $\sim$10\,Hz, demonstrating that language-based planning need not sacrifice latency for textual reasoning.}
\vspace{-1em}
\small
\resizebox{0.6\columnwidth}{!}{%
\begin{tabular}{@{}lccc@{}}
\toprule
\textbf{Model} & \textbf{Reasoning} & \textbf{Latency (ms)} & \textbf{Hardware} \\ 
\midrule
DriveVLM~\cite{tian2024drivevlm} & Yes & 410 & Orin X--2 \\
DriveGPT4-V2-8B~\cite{xu2025drivegpt4} & No & 2500 & --- \\
DriveGPT4-V2-1.5B~\cite{xu2025drivegpt4} & No & 345 & --- \\
DriveGPT4-V2-0.5B~\cite{xu2025drivegpt4} & No & 124 & --- \\
PlanTF~\cite{cheng2024plantf} & No & \textbf{12} & A6000 \\
DiffusionPlanner~\cite{zheng2025diffusionplanner} & No & 50 & A6000 \\
FlowPlanner~\cite{tan2025flow} & No & 83 & A6000 \\
\midrule
\lad & No & \underline{43} & A6000 \\
\lad (10 tokens max) & Yes & 102 & A6000 \\
\lad (40 tokens max) & Yes & 222 & A6000 \\
\bottomrule
\end{tabular}%
}
\label{tab:latency}
\vspace{-2em}
\end{table}

In example (a), \lad demonstrates cross-modal situational grounding: it interprets lane topology, pedestrian motion, and turn geometry directly from the scene context, enabling it to judge turn-safety and explain its reasoning.
Example (b) highlights relational reasoning, where \lad identifies spatial relationships between agents (e.g., which vehicle is behind or approaching) and uses these relations to anticipate potential conflicts.
Example (c) shows conditional and rule-aware reasoning: \lad evaluates counterfactuals, applies traffic laws such as pedestrian right-of-way, and understands when rules can be safely relaxed.

These capabilities translate directly into improved closed-loop behavior. \supplorinline{In the supplementary material, we present}{In Appendix~\ref{sec:supp_visual}, we present} additional visual comparisons showing \lad navigating complex intersections, roundabouts, and blocked-lane scenarios while articulating its high-level intent. In each case, the generated reasoning remains consistent with the scene layout and executed trajectory, and \lad outperforms PlanTF in scenarios requiring adaptive decision-making. \emph{We also provide video demonstrations for the interested reader.}

\vspace{-1em}
\section{Conclusion}
\vspace{-1em}
\label{sec:conclusion}

We presented two complementary approaches for autonomous driving planning that address different aspects of real-world complexity. \rad demonstrates that carefully designed rule-based planners remain highly competitive when equipped with richer topology exploration and goal-directed optimization, substantially improving geometric maneuvering capabilities while retaining the reliability and interpretability of structured planning. 

In parallel, \lad introduces the first real-time language-action planner for closed-loop driving. Through interruptible inference, \lad produces valid trajectories in a single forward pass while optionally generating language-based reasoning when compute permits, enabling semantic understanding of ambiguous traffic without sacrificing responsiveness required for safety-critical systems.

Together, these methods illustrate a practical path toward combining structured planning and foundation models in autonomous driving. Our results with \rad-\lad suggest that rules and language-grounded learning offer complementary capabilities, yielding systems that are both robust in routine driving and adaptable to the long tail of real-world scenarios.

\bibliographystyle{splncs04}
\bibliography{references}

\clearpage
\appendix

\setlength{\tabcolsep}{8pt}

\ifarxiv\else
\section*{Overview}

This supplementary material provides design rationale, ablations, and additional analysis supporting the main paper.
It is organized as follows: design rationale and extended discussion (Section~\ref{sec:faq}), \lad details and qualitative analysis (Section~\ref{sec:supp_lad}), \rad component ablations (Section~\ref{sec:supp_rad}), Val14 generalization results (Section~\ref{sec:supp_val14}), and training dataset descriptions (Section~\ref{sec:supp_datasets}).
\fi

\section{Design Rationale and Extended Discussion}
\label{sec:faq}

\noindent\textit{\textbf{Architecture \& Design Philosophy}}

\vspace{6pt}
\noindent
\textbf{Q: How does \lad's interruptible architecture enable reasoning before planning?}

\noindent
\textbf{A:}
Prior language-based planners couple action generation to full autoregressive text generation, creating a latency-quality tradeoff.
\lad's interruptible architecture avoids this: reasoning tokens precede the plan token, allowing the model to ``think before acting'' within a single forward pass.
Our ablations show that adding reasoning tokens improves planning quality, indicating that the plan token attends to its textual reasoning chain.
This design naturally extends to richer test-time reasoning strategies as future work, requiring no architectural changes.

\vspace{6pt}
\noindent
\textbf{Q: How is the reasoning budget controlled at inference?}

\noindent
\textbf{A:}
In the current evaluation, the token count is a fixed hyperparameter.
The key property is that the plan token \emph{always} produces a valid trajectory regardless of how many reasoning tokens precede it.
This makes the system directly compatible with external safety monitors that may demand immediate re-planning at any moment---the plan is never ``incomplete.''

\vspace{6pt}
\noindent
\textbf{Q: Why is a 0.6B LLM backbone sufficient, and what are the scaling prospects?}

\noindent
\textbf{A:}
Qwen3-0.6B was chosen deliberately to satisfy the real-time latency constraint while still demonstrating a core hypothesis: that language supervision improves planning.
The ablation in Table~\ref{tab:lad-ablations} confirms that language supervision consistently improves planning quality even at this scale.
Importantly, the \lad architecture (plan token, interruptible inference, adapters) is model-size agnostic; scaling to larger backbones requires no architectural changes, making studying scaling behavior a natural direction for future work.

\vspace{12pt}
\noindent\textit{\textbf{Context \& Positioning}}

\vspace{6pt}
\noindent
\textbf{Q: How does \lad relate to concurrent VLA/VLM planners?}

\noindent
\textbf{A:}
These methods~\cite{hwang2024emma, tian2024drivevlm, xu2025drivegpt4, zhou2025autovla} target a different operating regime: they process camera images, use larger backbones, and are primarily evaluated on vision-centric benchmarks or synthetic simulators.
\lad addresses a complementary setting, real-time language-action planning on nuPlan, the standard closed-loop benchmark. Accordingly, \lad is compared against the best performing methods on the nuPlan benchmark: \cite{cheng2024plantf, cheng2024pluto, zheng2025diffusionplanner, tan2025flow}.

\vspace{6pt}
\noindent
\textbf{Q: Why does \lad use vectorized inputs rather than camera or LiDAR inputs?}

\noindent
\textbf{A:}
End-to-end VLA/VLM planners~\cite{hwang2024emma, tian2024drivevlm, xu2025drivegpt4} study a different problem, jointly learning perception and planning from raw sensor data, and typically rely on open-loop evaluation, which does not reliably predict closed-loop performance~\cite{li2024ego, zhai2023rethinking}.
Our work addresses a complementary question: whether language supervision improves the \emph{planning} component itself. We evaluate in nuPlan's realistic closed-loop simulator, the standard setting for state-of-the-art planners~\cite{cheng2024plantf, cheng2024pluto, zheng2025diffusionplanner, tan2025flow}.
A key advantage of this setting is scalable language supervision: closed-loop simulators with logged data enable straightforward construction of trajectory-aligned QA pairs, as we demonstrate with DrivingQA and PlanningQA, leveraging behavior annotations~\cite{chang2025langtraj} as one source of grounding.

More broadly, we believe deployed planners will likely consume varying input configurations depending on available sensors and infrastructure, from vision and LiDAR alone to full stack inputs including HD maps and tracked agents.
Recent work in robotics~\cite{zitkovich2023rt, driess2023palm, black2025pi0} suggests that training across diverse modality combinations yields complementary gains.
Language supervision from log-replay simulators like nuPlan offers a particularly portable training signal, as it can be synthesized regardless of the underlying sensor modality.
\lad's modality-agnostic adapter pattern supports this direction: augmenting the encoder with, e.g., a vision backbone is feasible with few changes.

\begin{figure*}[t!]
    \centering
    \includegraphics[width=\linewidth]{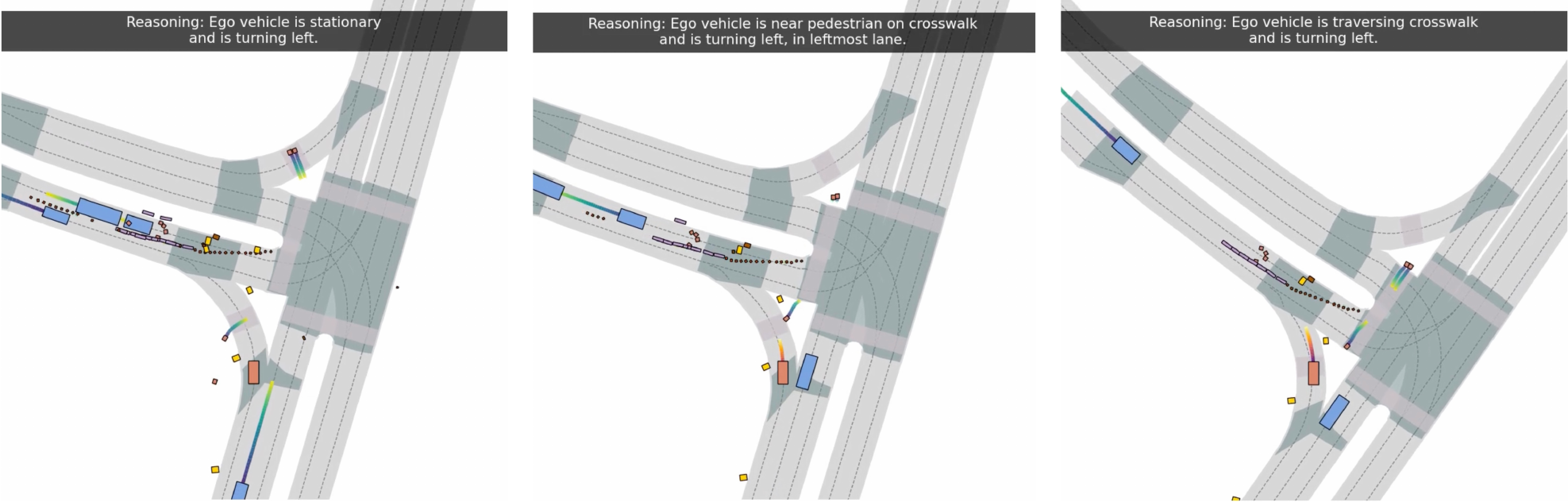}
    \caption{\lad demonstrates strong spatio-temporal understanding when navigating a complex left-turn. In each case, it correctly identifies nearby pedestrians, crosswalks, and turning geometry, and produces a safe, smooth trajectory while articulating its high-level intent (e.g., ``turning left,'' ``near pedestrian on crosswalk''). \lad's textual explanations remain consistent with the scene layout and the executed motion plan.}
    \label{fig:intent-1}
\end{figure*}

\vspace{12pt}
\noindent\textit{\textbf{Empirical Insights}}

\vspace{6pt}
\noindent
\textbf{Q: What is the key insight behind \rad's design?}

\noindent
\textbf{A:}
\rad contributes both a concrete algorithmic improvement, and the empirical finding it enables: these extensions close most of the gap between PDM-Closed and human performance on Test14-Hard, demonstrating that existing long-tail benchmarks predominantly capture capabilities absent from the baseline planner rather than intrinsic scenario difficulty.
The deliberate simplicity of \rad is what makes this insight useful: because the fix is simple, the large performance gain can be attributed directly to the new capabilities rather than to added model capacity or data.

\vspace{6pt}
\noindent
\textbf{Q: PDM-Closed already scores 15 proposals at every timestep. Why does it still need dynamic replanning?}

\noindent
\textbf{A:}
PDM-Closed is the strongest existing rule-based planner in the literature but was designed with a fixed topology. Despite scoring 15 candidates, PDM-Closed anchors all of them to proposal paths fixed at initialization, and there is no topological replanning after the first timestep.
If the ego drifts or paths become blocked, the underlying topology is never updated.
The official implementation reflects this design choice:

\begin{center}
{\fontsize{6.5pt}{7.5pt}\selectfont
\begin{BVerbatim}[frame=single,baselinestretch=0.95]
# L98 abstract_pdm_closed_planner.py
# TODO: Find additional conditions to trigger re-planning
create_new_proposals = self._iteration == 0
\end{BVerbatim}
}
\end{center}

\noindent
As a result, PDM-Closed does not support lane changes or off-route recovery. \rad addresses exactly this: it performs full topological replanning at every timestep, regenerating proposal paths from the current ego state and augmenting them with adjacent-lane centerlines.

\vspace{6pt}
\noindent
\textbf{Q: How do \rad and \lad complement each other?}

\noindent
\textbf{A:}
The two planners address different axes of difficulty.
Many Test14-Hard scenarios are geometric in character, which \rad already handles well, so the hybrid gain on this split is modest.
On InterPlan, which tests multi-agent interaction, the improvement from adding \lad is more pronounced, indicating that the language component contributes most in semantically complex scenarios. Beyond numeric scores, \lad provides \emph{interpretable reasoning} for every decision, which is a critical capability for deployment that pure rule-based systems cannot offer.

\vspace{6pt}
\noindent
\textbf{Q: Do the improvements generalize beyond long-tail benchmarks?}

\noindent
\textbf{A:}
Yes. As shown in Table~\ref{tab:val14}, \rad marginally improves over PDM-Closed on Val14, confirming that lane-change capability and dynamic replanning do not harm general driving.
Among imitation-based learned planners, \lad outperforms PlanTF on Val14 as well.
Notably, \rad's trajectory vocabulary is drawn from the full nuPlan training set, so its strong Test14-Hard performance does not reflect overfitting to rare maneuvers.
The human-gap analysis in Table~\ref{tab:human_gap} reinforces this: both planners remain close to human performance on Val14, yet PDM-Closed's gap on Test14-Hard is roughly 4$\times$ larger than \rad's. This insight indicates that most of the ``difficulty'' captured by Test14-Hard reflects capabilities outside PDM-Closed's design scope, while \rad's improvements generalize across both standard and long-tailed conditions.

\begin{figure*}[t!]
    \centering
    \includegraphics[width=\linewidth]{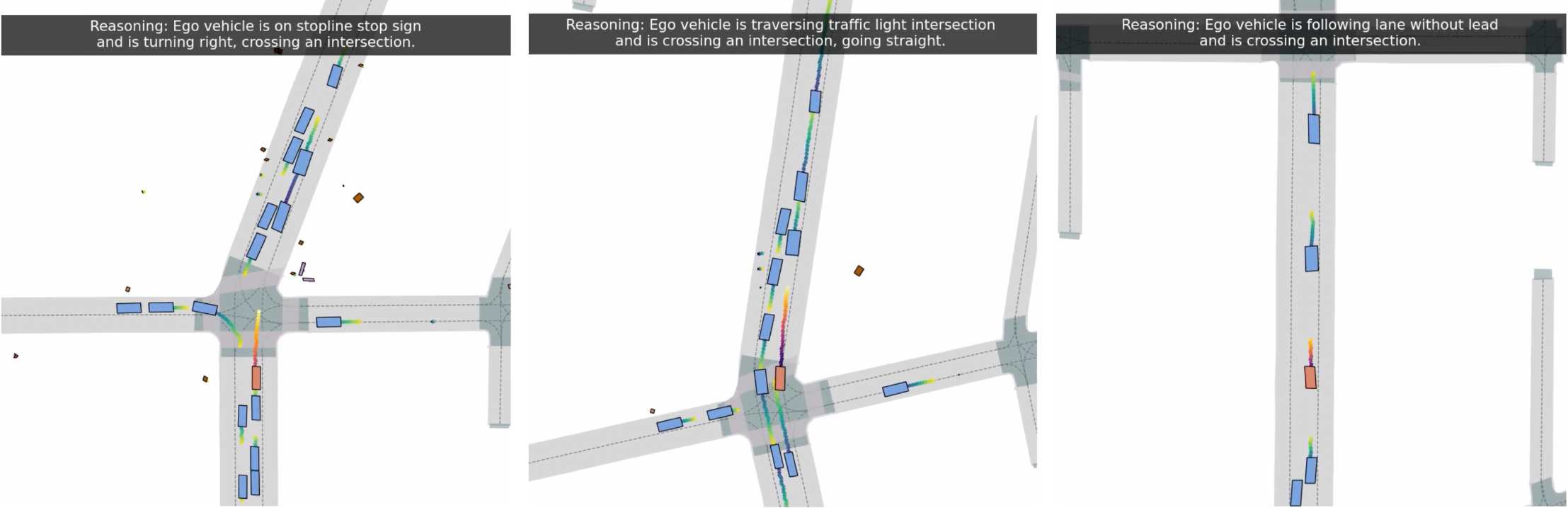}
    \caption{\lad handles diverse road configurations by accurately understanding lane topology, control rules, and surrounding agents. Its generated reasoning reflects this situational awareness (e.g., ``on stopline stop sign,'' ``crossing an intersection'', ``following lane without lead''), and its trajectory choices align with these high-level descriptions.}
    \label{fig:intent-2}
\end{figure*}

\vspace{12pt}
\noindent\textit{\textbf{Reproducibility}}

\vspace{6pt}
\noindent
\textbf{Q: What is the latency measurement methodology?}

\noindent
\textbf{A:}
Prior work has noted that LLM-based planners face ``significant challenges, including elevated resource consumption and extended inference times, which pose substantial obstacles to practical deployment''~\cite{chen2024asyncdriver}.
All closed-loop planners in Table~\ref{tab:latency} (PlanTF, DiffusionPlanner, FlowPlanner, and \lad) are measured on the same A6000 hardware under identical evaluation conditions.
Our results show that this limitation is not fundamental: \lad is \emph{fast enough for closed-loop deployment} (10\,Hz with reasoning, 20\,Hz without), and among closed-loop planners evaluated on this hardware, \lad is competitive while being the only method that also produces interpretable reasoning.
DriveVLM's reported Orin latency is included only for reference to concurrent language-based planners.

\vspace{6pt}
\noindent
\textbf{Q: Will code, models, and data be released?}

\noindent
\textbf{A:}
Yes. We will release code and model weights.

\section{\lad: Language Based Autonomous Driving}
\label{sec:supp_lad}

\subsection{Structured Reasoning for Inference}
\label{sec:supp_structured_reasoning}

To satisfy strict latency constraints while retaining interpretability, \lad uses \emph{structured reasoning templates} that restrict generation to task-relevant tokens. Instead of open-ended text generation for the planning reasoning, we use templates with designated fill-in fields, 

\texttt{Ego is \{scenario\_type\} and is \{meta\_action\}}\tok{plan}.

\noindent
The model generates only the tokens within curly braces (e.g., meta actions like \texttt{following\_lane}, \texttt{turning\_left}). Do note that the DrivingQA dataset is open ended and is part of the data-mix while training.

\paragraph{Training Augmentation.}
During training, we randomly truncate reasoning mid-generation before the \tok{plan} token. This augmentation ensures the model learns to produce valid plans regardless of how much reasoning context is available. We observed no drop in planning performance from this truncation strategy.

\paragraph{Multi-Turn Capability.}
While \lad is trained exclusively on single-turn QA pairs from Interaction-QA dataset, it retains the multi-turn conversational ability of its base language model (Qwen3-0.6B). The examples in Figure~\ref{fig:chatbot-examples} demonstrate this emergent capability at inference time.

\subsection{Latency Measurement Methodology}
\label{sec:supp_latency}

Our latency measurements reported in Table~\ref{tab:latency} represent \emph{end-to-end inference time}, measured from receiving vectorized inputs (map elements, agent states) to extracting the final waypoint trajectory. This includes the PlanTF encoder forward pass, MLP adapter projection, language model prefill (and optional autoregressive decoding for reasoning tokens), planning head forward pass, and trajectory vocabulary lookup. Measurements are averaged over 1000 warm-start inference calls. The reported times do not include data loading or raw sensor pre-processing.

\subsection{More Visual Comparisons}
\label{sec:supp_visual}

\begin{figure*}[t!]
    \centering
    \includegraphics[width=\linewidth]{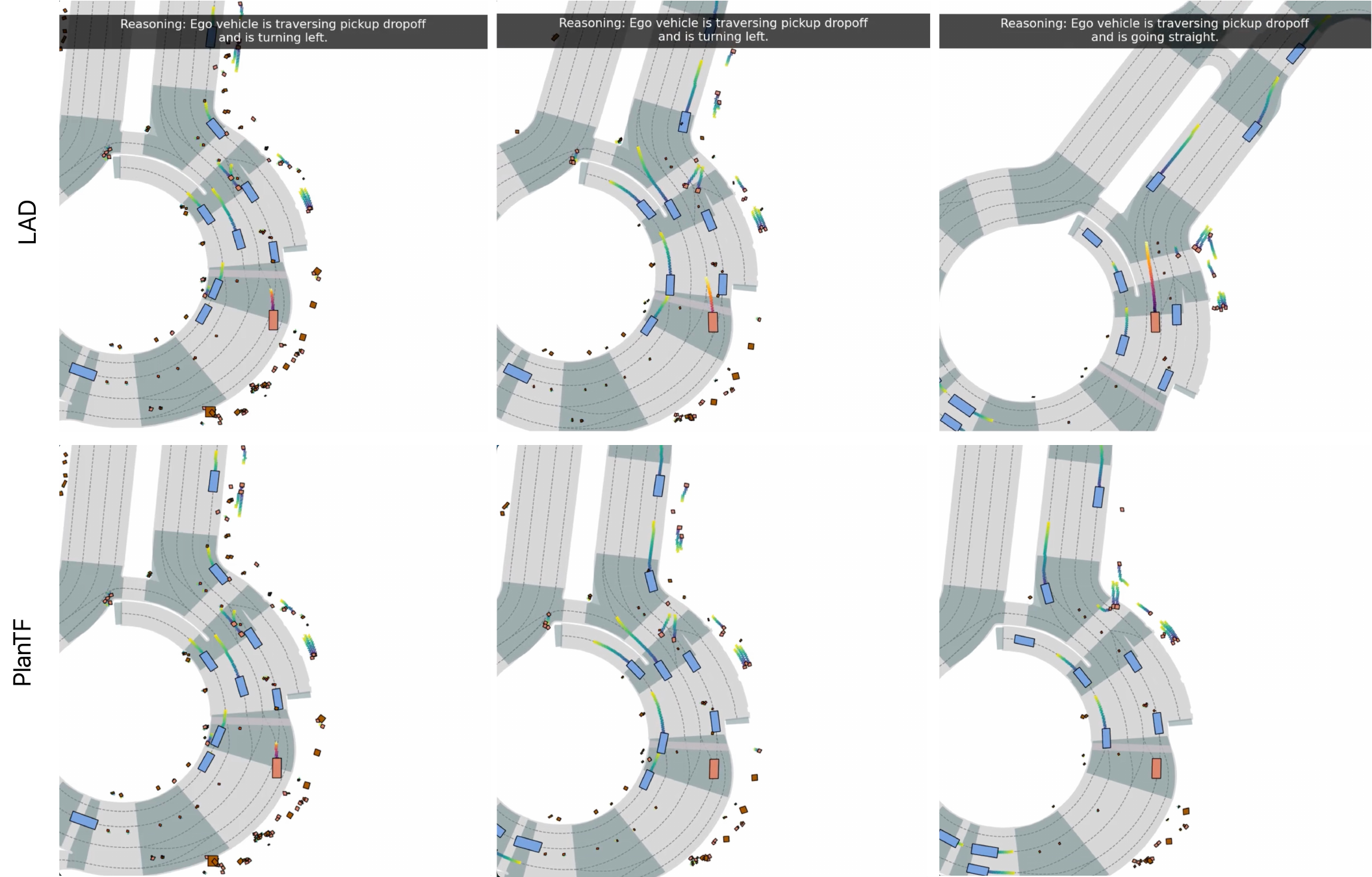}
    \caption{\lad produces safe and consistent trajectories through this complex roundabout while simultaneously articulating its high-level intent (e.g., ``ego vehicle is traversing pickup-dropoff and is turning left''). In contrast, PlanTF~\cite{cheng2024plantf} frequently hesitates or commits to suboptimal maneuvers. \lad's textual reasoning aligns with its chosen motion plan, providing interpretable justification for each action.}
    \label{fig:plantf-1}
\end{figure*}

\begin{figure*}[t!]
    \centering
    \includegraphics[width=\linewidth]{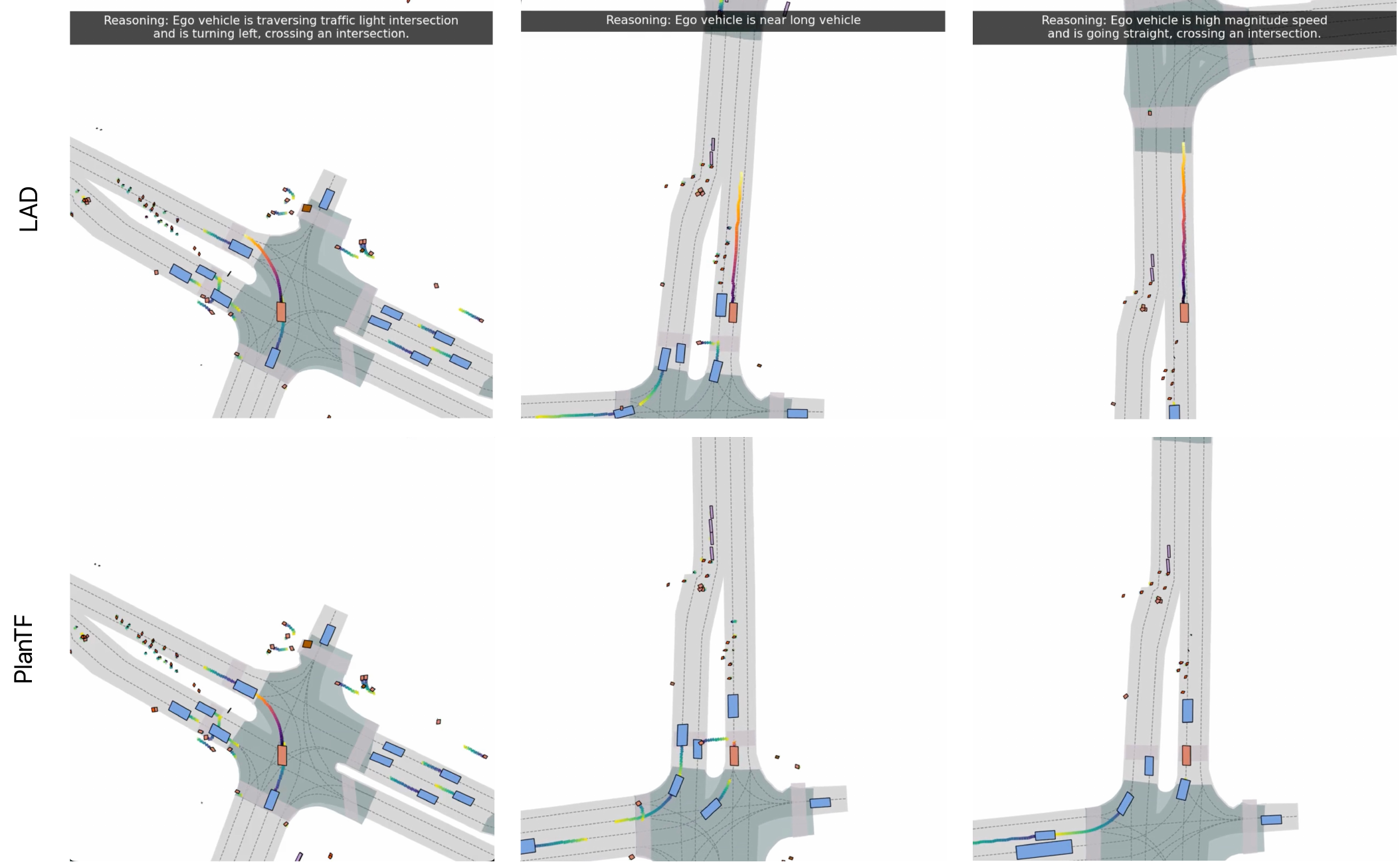}
    \caption{
        \lad correctly infers from the scene that the ego lane is blocked by static objects. It then selects the farther unblocked lane and executes a safe maneuver through the intersection, identifying relevant dynamic objects (``long vehicle''). In contrast, PlanTF~\cite{cheng2024plantf} continues to follow the blocked lane or hesitates, failing to account for the static obstacles.
    }
    \label{fig:plantf-2}
\end{figure*}

We present a few scenarios where \lad shows good spatio-temporal understanding of the map and other agents around while navigating (See Figure~\ref{fig:intent-1} and Figure~\ref{fig:intent-2}). 
We also present some visual comparisons between \lad and PlanTF~\cite{cheng2024plantf} in Figure~\ref{fig:plantf-1} and Figure~\ref{fig:plantf-2}. 
We also show more visual results in our associated supplementary video.

\section{\rad Ablation Studies}
\label{sec:supp_rad}

In this section, we provide ablation studies of the various algorithmic novelties relative to PDM-Closed. The effectiveness of these components is validated on subsets of nuPlan Test14 and InterPlan subsets that do not overlap with the benchmark subsets.



\begin{table}[h]
\centering
\caption{\small \rad Ablation on nuPlan Test14 subset}
\vspace{-1em}
\begin{tabular}{cccc}
\toprule
\textbf{PDM-Closed} & \textbf{+Replan} & \textbf{+Vocab Aug.} & \textbf{+Adj. Centerlines}\\
\midrule
92.75 & 93.92 & 94.40 & 94.51\\
\bottomrule
\end{tabular}
\label{tab:rad_replan}
\vspace{-3em}
\end{table}

\begin{table}[h]
\centering
\caption{\small \rad Ablation on InterPlan subset}
\vspace{-1em}
\begin{tabular}{ccc}
\toprule
\textbf{\rad} & \textbf{+Goal Opt.}  & \textbf{+Rule Relaxation}\\
\midrule
 84.31 & 90.22 & 92.99\\ 
\bottomrule
\end{tabular}
\label{tab:rad_goal}
\end{table}

\section{Additional Results}
\label{sec:supp_val14}

\subsection{Results on Regular Driving Situations} 

Table~\ref{tab:val14} reports performance on nuPlan Val14~\cite{caesar2021nuplan}, the standard split representative of normal driving situations. \rad marginally improves over PDM-Closed, confirming that dynamic replanning and lane-change capability do not degrade nominal driving quality. Among imitation-based learned planners, \lad outperforms PlanTF, demonstrating that language supervision does not harm normal driving performance.

\setlength{\tabcolsep}{8pt}

\begin{table}[t]
\centering
\caption{\small \textbf{Val14 (Reactive) results.} \rad marginally improves over PDM-Closed, confirming that dynamic replanning and lane-change capability do not degrade nominal driving quality. \lad outperforms PlanTF among learned planners, demonstrating that language supervision preserves normal driving performance.}
\small
\begin{tabular}{llc}
\toprule
\textbf{Type} & \textbf{Planner} & \textbf{Val14 (R)} \\
\midrule
\color{gray}{Expert} & \color{gray}{Log Replay} & \color{gray}{93.68} \\
\midrule
\multirow{3}{*}{Rule}
 & IDM~\cite{treiber2000congested} & 79.31 \\
 & PDM-Closed~\cite{dauner2023pdmclosed} & \underline{92.12} \\
 & \textbf{\rad\ (Ours)} & \textbf{92.31} \\
\midrule
\multirow{5}{*}{Learned}
 & PlanTF~\cite{cheng2024plantf} & 77.07 \\
 & PLUTO~\cite{cheng2024pluto} & 80.01 \\
 & DiffusionPlanner~\cite{zheng2025diffusionplanner} & \underline{82.80} \\
 & FlowPlanner~\cite{tan2025flow} & \textbf{83.31} \\
 & \textbf{\lad\ (Ours)} & 78.40 \\
\midrule
\multirow{5}{*}{Hybrid}
 & PLUTO~\cite{cheng2024pluto} & 87.00 \\
 & STR2-CKS-800m~\cite{sun2024generalizingmotionplannersmixture} & 92.12 \\
 & DiffusionPlanner~\cite{zheng2025diffusionplanner} & \textbf{92.90} \\
 & FlowPlanner~\cite{tan2025flow} & \underline{92.38} \\
 & \textbf{\rad-\lad\ (Ours)} & 92.35 \\
\bottomrule
\end{tabular}

\label{tab:val14}
\end{table}

\begin{table}[t]
\centering
\caption{\small \textbf{Impact of Downstream Controller.} Replacing the default LQR controller with iLQR improves \lad's score by 4.07, consistent with the hidden imitation gap arising due to not accounting for downstream controller~\cite{cheng2024plantf}. Thus, controller-awareness is an important consideration for trajectory-based planners.}
\small
\begin{tabular}{llc}
\toprule
\textbf{Planner} & \textbf{Controller} & \textbf{Val14 (R)} \\
\midrule
\lad\ & LQR & 78.40 \\
\lad\ & iLQR & 82.47 \\
\bottomrule
\end{tabular}
\label{tab:lqr-ilqr-tracking}
\end{table}

\subsection{Impact of Downstream Controller}
\label{sec:controller_impact}

Prior work~\cite{cheng2024plantf} suggested the existence of a hidden imitation gap which arises due to the discrepancy between trajectory-based planner and the downstream controller. The expert trajectory serves as the ground truth during the training of the imitation-based planner. The predicted trajectory is processed by a downstream controller (could be LQR or iterative LQR or some other algorithm) and the underlying system dynamics, which are not considered during training. Thus, during roll out in closed-loop evaluation, this discrepancy may lead to a decrease in planning performance as predictions do not consider what can be actually actuated or tracked by the downstream controller.

In Table~\ref{tab:lqr-ilqr-tracking}, we confirm this hidden imitation gap is the reason impeding \lad's performance in normal driving situations. Changing the downstream controller from LQR to a more robust iterative LQR (iLQR) controller improves performance by 4.07\% closing the gap between \lad and other state-of-the-art imitation methods~\cite{zheng2025diffusionplanner, tan2025flow}. 

However, we note that these iLQR results depart from the standard nuPlan evaluation protocol\footnote{\url{https://nuplan-devkit.readthedocs.io/en/latest/competition.html}}, which couples a fixed LQR controller with the simulation. As our results show, this coupling is suboptimal: the LQR controller degrades the closed-loop performance of trajectory-based planners. Prior work~\cite{jaeger2025carl} corroborates this finding, showing that even the human expert trajectory loses $\sim$3 points under the default LQR controller compared to a more accurate iLQR controller. This suggests that nuPlan scores partially reflect controller quality rather than planning quality alone.

We concur with prior work~\cite{jaeger2025carl} that simulators and benchmarks for autonomous driving should decouple control from simulation. The choice of controller is inherently tied to the planner's output representation, an observation that has long motivated end-to-end approaches to autonomous driving~\cite{bojarski2016end, prakash2021multi, hu2023planning}. Bridging this gap is an orthogonal research direction: our work focuses on improving \emph{planning} through language supervision. Addressing the control interface, whether through alternative output representations~\cite{jaeger2025carl}, learning the controller~\cite{cheng2024plantf}, smoother action execution strategies~\cite{black2025real}, or benchmark redesign, is a complementary but distinct line of work.

\subsection{Does \rad overfit to long-tailed situations?} 

Table~\ref{tab:human_gap} compares the performance gap between planners and human expert driving on Val14 versus Test14-Hard. While both RAD and PDM-Closed achieve near-human performance on Val14, PDM-Closed exhibits a notably larger gap on the long-tailed splits.

Crucially, \rad's improvements stem from general-purpose driving capabilities, i.e., lane changes, dynamic topology replanning, and goal-directed optimization, rather than heuristics tailored to specific failure modes.

The asymmetry in human gaps is revealing. Both methods are near-human on Val14, so baseline planning quality is comparable. On Test14-Hard, however, PDM-Closed's gap widens to roughly $2\times$ \rad's gap. Since \rad differs from PDM-Closed only in structural capabilities, this excess gap indicates that many Test14-Hard scenarios are ``hard'' not because of inherent complexity, but because PDM-Closed lacked certain structural capabilities.

Finally, \rad's Val14 score marginally \emph{improves} over PDM-Closed, ruling out the overfitting hypothesis. If \rad were specialized to long-tail scenarios at the expense of normal driving, we would expect a regression on Val14. Instead, the consistent or improved performance across both splits confirms that \rad's gains are attributable to improvements in general planning capabilities.

\setlength{\tabcolsep}{8pt}

\begin{table}[t]
\centering
\caption{\small \textbf{Human-gap analysis across splits.} Both \rad and PDM-Closed achieve near-human performance on Val14, but \rad maintains a significantly smaller gap on Test14-Hard, suggesting that \rad's lane-change and replanning capabilities generalize to long-tail scenarios without overfitting.}
\small
\begin{tabular}{lcccc}
\toprule
& \multicolumn{2}{c}{\textbf{Val14}} & \multicolumn{2}{c}{\textbf{Test14-Hard}} \\
\cmidrule(lr){2-3} \cmidrule(lr){4-5}
\textbf{Method} & \textbf{Score} & \textbf{Human Gap} & \textbf{Score} & \textbf{Human Gap} \\
\midrule
PDM-Closed & 93.20 & 0.80 & 75.19 & 10.77 \\
\rad & 92.31 & 1.69 & 80.53 & 5.43 \\
Human (Expert) & 94.00 & -- & 85.96 & -- \\
\bottomrule
\end{tabular}
\label{tab:human_gap}
\end{table}

\section{Datasets}
\label{sec:supp_datasets}

\subsection{DrivingQA Dataset}
\label{sec:supp_interaction_qa}

Training multimodal language models for autonomous driving requires grounded question-answering data that captures the nuanced dynamics of multi-agent interactions. 
Existing driving QA datasets often focus on object recognition or simple scene descriptions, lacking structured annotations for ego-centric planning decisions and inter-agent relationships.
To address this gap, we introduce \textbf{DrivingQA}, a synthetically generated instruction-tuning dataset built on top of NuPlan~\cite{caesar2021nuplan}.
DrivingQA contains 1.2 million question-answer pairs spanning 3.3 million driving scenarios taken from 8,457
nuplan scenes, with explicit annotations for ego vehicle plans, agent-of-interest (AOI) behaviors, and multi-agent interactions.

\paragraph{Data Sources.}
We leverage heuristic and human annotated behavior annotations from InterDrive~\cite{chang2025langtraj} which provide structured per-agent behavior labels (e.g., lane position, turn intent, speed state, intersection behavior) for nuPlan scenes (a scene is a 20 second driving log). We additionally employ GPT-4o rephrased descriptions that introduce natural language diversity for the same underlying behaviors. We associate each scene with one or more NuPlan scenario types (from a taxonomy of types such as \texttt{starting\_left\_turn}), along with precise temporal annotations indicating when each \textit{scenario} occurs within the 20-second scene (\texttt{min\_time}, \texttt{max\_time} in seconds).


\paragraph{Generation Pipeline.}
For each scene with annotated interactions, we construct structured prompts for three subject categories: the \emph{ego vehicle}, the \emph{agent of interest} (the primary interacting agent), and \emph{other agents} present in the scene.
Each prompt includes the subject's behavior description, interaction context, and scenario type metadata. We employ Qwen2.5-32B~\cite{qwen25vl} to generate 2 to 4 QA pairs per ego/agent of interest subject and 1 to 3 pairs for every other agent.

To ensure \emph{entity-agnostic} generalization, all agents are referenced using special tokens: \tok{ego} for the ego vehicle, \tok{agent_of_interest} for the primary interacting agent, and \tok{agent} for all other agents. While training, we replace these tokens with the numerical ID's assigned to each agent. Each answer is tagged with a provenance label, either \textit{stated} (directly from annotations), \textit{deduced} (inferred from context), or \textit{unknown} (insufficient evidence), which enables us to perform filtering before training.

\paragraph{Question-Answer Diversity.} The system prompt encourages lexical and structural diversity, varying interrogatives (what/which/how/why/does/can), paraphrasing answers, and including negative/counterfactual questions (e.g., ``Is Agent \tok{ego} turning left?'' when the ground truth is going straight). Questions span multiple categories including \emph{planning} (ego maneuver intent), \emph{behavior} (agent motion states), \emph{interaction} (yielding, priority, relative positioning), and \emph{spatial} (lane occupancy, intersection traversal).

\subsection{PlanningQA Dataset}
\label{sec:supp_planningqa}

\begin{figure*}[!htbp]
{
\centering
\begin{tcolorbox}[
    colback=teal!5,
    colframe=teal!70,
    title={\textbf{System Directive}},
    fonttitle=\small\bfseries,
    boxrule=0.5pt,
    arc=2pt,
    left=4pt, right=4pt, top=0.5pt, bottom=0.5pt
]
\footnotesize\ttfamily
You generate high-quality instruction-tuning QA data for self-driving scenarios. Assume your role is of a Planner agent reasoning about the scene and the agents in it.\\[1pt]
\textbf{Entity References:} In all questions and answers, use these exact references:\\
\hspace*{1em}-- ego vehicle: `Agent <|ego|>'\\
\hspace*{1em}-- primary interacting agent: `Agent <|agent\_of\_interest|>'\\
\hspace*{1em}-- any other agent: `Agent <|agent|>'\\
Never use raw IDs. IDs belong only in metadata.\\[1pt]
\textbf{Scene Context:} A scene is a 20-second driving segment. Scenarios are triggered by ego behavior at specific time instants. Each scenario has a type from the NuPlan taxonomy.\\[1pt]
\textbf{Grounding Rules:} Only use facts in the provided JSON. If a detail is not stated and cannot be deduced, answer `Unknown' with answer\_source `unknown'. Do not refer to the input text in your answers.\\[1pt]
\textbf{Diversity:} Vary interrogatives (what/which/how/why/does/can), paraphrase answers, include negative/counterfactual questions. Avoid repeating the same template.\\[1pt]
\textbf{Output:} Valid JSON array with schema: \{question, answer, answer\_source, tags\}.
\end{tcolorbox}

\vspace{0.05em}

\begin{tcolorbox}[
    colback=cyan!5,
    colframe=cyan!60!teal,
    title={\textbf{User Prompt (per-subject)}},
    fonttitle=\small\bfseries,
    boxrule=0.5pt,
    arc=2pt,
    left=4pt, right=4pt, top=0.5pt, bottom=0.5pt
]
\footnotesize\ttfamily
\textbf{Task:} Produce 2 to 4 QA items for the \textbf{Ego} subject.\\[1pt]
\textbf{Additional Instructions:} Ask about current state, scenario type, plan, interactions with agent of interest.\\[1pt]
\textbf{Scene token:} \{scene\_token\}\\[1pt]
\textbf{Scenario type aggregates}: (context only, do NOT ask about min\_time/max\_time)\\
\hspace*{1em}[\{"scenario\_type": "on\_intersection", "agent\_track\_token": null,\\
\hspace*{2em}"min\_time": 8.15, "max\_time": 9.65\},\\
\hspace*{1em} \{"scenario\_type": "traversing\_intersection", ...\}]\\[1pt]
\textbf{Subject:} ego, Type: VEHICLE\\
\textbf{Has interaction}: true\\[1pt]
\textbf{Input data (JSON)}:\\
\{"reasoning\_text": "Is in middle lane, crossing an intersection, going straight.",\\
\hspace*{1em}"ego": \{"other\_agent\_token": "<|nuplan\_token|>"\}, "has\_interaction": true\}
\end{tcolorbox}

\vspace{0.3em}

\begin{tcolorbox}[
    colback=green!5,
    colframe=green!60!black,
    title={\textbf{Output Schema}},
    fonttitle=\small\bfseries,
    boxrule=0.5pt,
    arc=2pt,
    left=4pt, right=4pt, top=0.5pt, bottom=0.5pt
]
\footnotesize\ttfamily
[\{"question": "What maneuver is Agent <|ego|> planning?",\\
\hspace*{1em}"answer": "Go straight through the intersection.",\\
\hspace*{1em}"answer\_source": "stated",\\
\hspace*{1em}"tags": ["planning", "intersection"]\}, ...]
\end{tcolorbox}
}
\vspace{-1em}
\caption{\textbf{DrivingQA generation prompt.} Template used to synthesize trajectory-grounded QA pairs from driving logs, for \lad's scalable language supervision pipeline.}
\label{fig:interaction_qa_prompt}
\end{figure*}

While DrivingQA (Section~\ref{sec:supp_interaction_qa}) provides broad scene understanding through open-ended question-answer pairs, it lacks temporal grounding to the ego vehicle's immediate action. PlanningQA addresses this gap by pairing each training trajectory with a short textual description of the ego's behavior, temporally aligned with the ground-truth waypoints. This provides the model with action-conditioned language supervision: text that describes \emph{what the ego is doing when the plan is executed}, rather than general scene-level information.

\paragraph{Scenario Type.}
Each nuPlan scenario carries a \emph{scenario type} label drawn from nuPlan's official taxonomy of 75 types~\cite{caesar2021nuplan}.
These types are algorithmically mined from driving logs via atomic event primitives (e.g., intersection entry, high lateral acceleration, dense traffic) and cover both frequent maneuvers (e.g., \texttt{following\_\\lane\_with\_lead}, \texttt{starting\_left\_turn}, \texttt{changing\_lane}) and rare long-tail events (e.g., \texttt{near\_miss}, \texttt{waiting\_for\_pedestrian\_to\_cross}, \texttt{traversing\_pickup\_\\dropoff}).
Because nuPlan natively provides temporal bounds (\texttt{min\_time}, \texttt{max\_\\time}) for each scenario within a scene, we can associate the correct scenario type, ensuring alignment between the label and the ground-truth trajectory.

\paragraph{Meta-Action.}
In addition to the scenario type, we annotate each planning timestep with a \emph{meta-action} label describing the ego vehicle's high-level behavioral intent.
These labels are derived from the heuristic single-agent behavior annotations provided by InterDrive~\cite{chang2025langtraj}, which assigns structured per-agent labels (e.g., lane position, turn intent, speed state, intersection behavior) based on calibrated geometric and kinematic heuristics applied to the nuPlan driving logs.
We extract and map the ego-relevant subset of these annotations to a compact set of meta-actions such as \texttt{following\_lane}, \texttt{turning\_left}, \texttt{turning\_right}, \texttt{stationary}, and \texttt{lane\_change}.

\paragraph{Template Format.}
During training, PlanningQA supervision is provided as a short structured sentence preceding the \tok{plan} token. Because the template is short and fixed-format, it adds minimal latency during inference while providing action-aligned textual supervision that is complementary to the open-ended DrivingQA pairs.

\end{document}